\documentclass[10pt,journal,compsoc]{IEEEtran}
\newif\ifpeerreview

\peerreviewfalse

\usepackage[nocompress]{cite}
\usepackage{url}
\usepackage{amsmath,amssymb,graphicx}

\usepackage{times}
\usepackage{epsfig}
\usepackage{graphicx}
\usepackage{graphics}
\usepackage{amsmath}
\usepackage{amssymb}
\usepackage[linesnumbered,ruled,vlined]{algorithm2e}
\usepackage{booktabs} %
\usepackage{overpic}
\usepackage{tikz}
\usepackage{bbm}
\usepackage{bm}
\usepackage{amsbsy}
\usepackage{multirow}
\usepackage{nicefrac}
\usepackage[pagebackref=true,breaklinks=true,colorlinks,bookmarks=false]{hyperref}

\definecolor{red1}{rgb}{0.8431    0.1882    0.1529}
\definecolor{orange1}{rgb}{0.9882    0.5529    0.3490}
\definecolor{yellow1}{rgb}{0.9961    0.8784    0.5451}
\definecolor{blue1}{rgb}{0.3000    0.3000    0.8490}
\definecolor{green2}{rgb}{0.2    0.8    0.3}
\definecolor{green1}{rgb}{0.1020    0.5961    0.3137}
\definecolor{red2}{rgb}{0.8431    0.1882    0.1529}

\newcommand{\av}{\mathbf{a}}

\newcommand{\uv}{\mathbf{u}}
\newcommand{\vv}{\mathbf{v}}
\newcommand{\wv}{\mathbf{w}}
\newcommand{\phiv}{\boldsymbol{\phi}}

\usepackage[switch]{lineno}

\newcommand{\paperID}{057}

\title{Projected Distribution Loss\\ for Image Enhancement}

\author{Mauricio Delbracio, Hossein Talebei and Peyman Milanfar%
\IEEEcompsocitemizethanks{\IEEEcompsocthanksitem The authors are with Google Research, Mountain View, CA 94043 USA.
The first two authors contributed equally to this work.\protect\\
}%
}

\begin{document}

\IEEEtitleabstractindextext{%
\begin{abstract}
Features obtained from object recognition CNNs have been widely used for measuring perceptual similarities between images. Such differentiable metrics can be used as perceptual learning losses to train image enhancement models. However, the choice of the distance function between input and target features may have a consequential impact on the performance of the trained model. While using the norm of the difference between extracted features leads to limited hallucination of details, measuring the distance between distributions of features may generate more textures; yet also more unrealistic details and artifacts. In this paper, we demonstrate that aggregating 1D-Wasserstein distances between CNN activations is more reliable than the existing approaches, and it can significantly improve the perceptual performance of enhancement models. More explicitly, we show that in imaging applications such as denoising, super-resolution, demosaicing, deblurring and JPEG artifact removal, the proposed learning loss outperforms the current state-of-the-art on reference-based perceptual losses. This means that the proposed learning loss can be plugged into different imaging frameworks and produce perceptually realistic results. 
\end{abstract}
}

\maketitle
\thispagestyle{empty}

\IEEEraisesectionheading{
  \section{Introduction}\label{sec:introduction}
}

\IEEEPARstart{I}{mage} restoration has seen remarkable  progress in recent years mostly coming hand-in-hand with the success of deep neural networks. Greater computational power, stable and accessible training frameworks as well as a large amount of data have enabled deep image processing models that exceed or are on par with those conceived through careful and artisan modeling.

However, how to train deep models in such a way that the restored images capture the realism of natural images remains an open challenge. The most popular approach for training deep image models is through supervised learning in which a loss function measuring the difference with respect to a reference or ground-truth image is minimized~\cite{zhao2016loss,lim2017enhanced,tao2018scale,chen2018learning}.

Perhaps the most common approach is to use a loss function that measures the difference directly between pixel values by some standard norm (e.g., $L_1$ or $L_2$). The pixel loss suffers from the well-known problem of regression to the mean. Since inverse problems are generally poorly conditioned, there are countless possible explanations for a given observed image. Minimizing a $L_2$ pixel loss leads to predicting an average image that generally looks blurry and lacks of details (e.g., grain, noise, edge contrast).

Generative adversarial networks~\cite{goodfellow2014generative,arjovsky2017wasserstein}, and in particular adversarial losses~\cite{ledig2017photo,isola2017image,kupyn2018deblurgan,kupyn2019deblurgan} are among recent approaches that lead to more realistic images. However, these networks are generally hard to train, due to a min-max type optimization~\cite{arora2017generalization,salimans2016improved,arjovsky2017wasserstein}. Additionally, since GANs are trained to minimize the distance to the manifold of natural images, they generally introduce significant hallucinations: what is perceived as real content for one image may be seen as hallucination in a different image~\cite{cohen2018distribution}.

\begin{figure}[!t]
\vspace{-0 mm}
\begin{center}
\includegraphics*[viewport=1 1 440 350,scale=0.55]{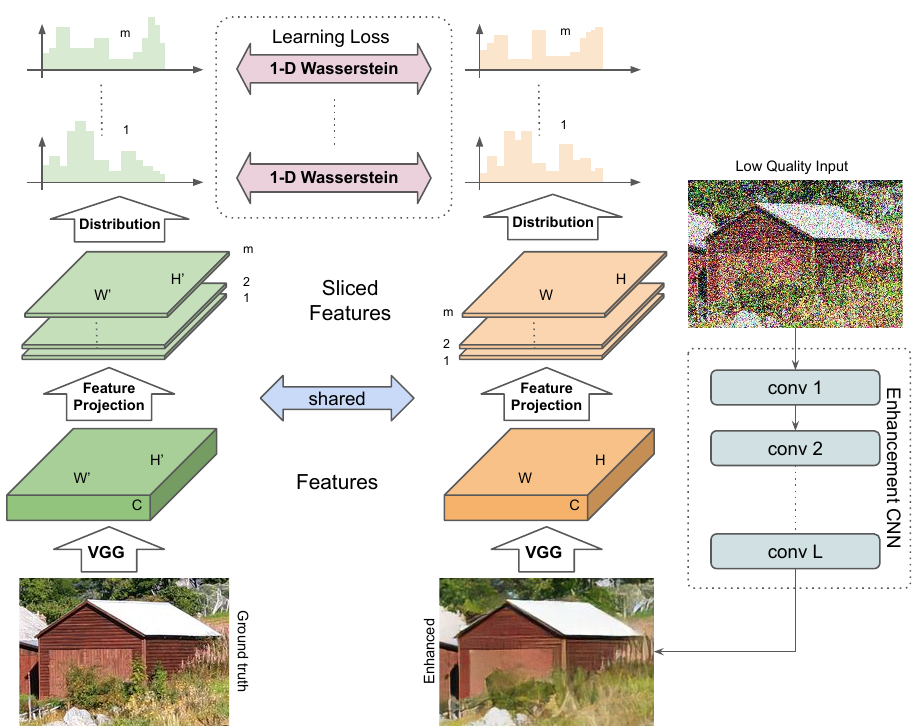}
\end{center}
\vspace{-2 mm}
{\caption{The proposed image enhancement framework. Our enhancement CNNs are trained with a loss that measures 1-D Wasserstein distance between distributions of the VGG features~\cite{simonyan2014very}.} \label{fig:framework}}
\vspace{-4 mm}
\end{figure}

Johnson et al.~\cite{johnson2016perceptual} showed that networks trained for image recognition tasks produce deep features (i.e., low / mid / high level image representations) that can capture perceptual information well. Based on this powerful observation, the authors proposed to augment a pixel loss (or replace it) by penalizing the $L_1/L_2$ difference in a deep feature space. The perceptual loss leads to images with more details and improve the missing realism ingredient. This has been recently analyzed in~\cite{zhang2018unreasonable,tarig2020why} and further exploited in the contextual loss ~\cite{mechrez2018contextual,mechrez2018maintaining} and contextual bilateral loss~\cite{zhang2019zoom}.

In this work, we build on recent findings and introduce a new loss function that measures differences in the feature space by comparing the distributions of an image and its reference counterpart. 
This allows recovering details (texture/grain) without forcing them to be located in the exact same spatial position as in the reference image. This is a critical flexibility for tackling inverse imaging problems. For instance, one can consider the problem of single image super-resolution where a low resolution image can be up-scaled to slightly different high-resolution images with identical feature distributions. This implies that given the low resolution observation, there are multiple high-resolution images that lead to the low-resolution observation. Any pixel loss or $L_1$ distance on features (Johnson et al.~\cite{johnson2016perceptual}) learns to predict the mean target (i.e., conditional expectation given the low resolution observation), leading to a less realistic (and often blurry) image.
On the other hand since we compare an image directly with its reference, the amount of hallucinated content is better controlled than in an adversarial loss.

Our proposed loss function has some similarity to style transfer works~\cite{gatys2016image}. In neural style transfer, a convolutional network is used to merge the content of one guide image with the style of another one. A neural network is trained for each different input style by using two different and complementary losses. The content loss is defined as the element-wise distance between feature maps of both images, while style loss penalizes difference in the statistics of the feature maps~\cite{gatys2016image,li2017demystifying,li2017universal}.

Our distribution loss makes explicit use of the fact that the extracted features belong to a metric space and therefore their geometries as well as their distributions are both relevant. This leads us to pose the problem as one of optimal transport~\cite{villani2008optimal}, in which we seek to minimize the cost of transforming the input feature distribution into the reference one.  The optimal transport problem and the Wasserstein distance\cite{villani2008optimal,peyre2019computational} provide an elegant and formal way to measure distance between distributions by transporting one distribution into another.  However, directly minimizing the Wasserstein distance on a high-dimensional distribution is in general intractable~\cite{cuturi2013sinkhorn}. Rabin et al.~\cite{rabin2011wasserstein} introduced the sliced-Wasserstein distance that makes explicit use of the closed-form solution in one-dimension as an elegant tractable approximation. The sliced variant is based on the invertibility of the Radon transform, and consists of projecting the data samples and computing an average one-dimensional distance between the marginal distributions at all possible orientations. Our loss function for comparing feature distributions is in effect an implementation of the sliced Wasserstein loss where we aggregate 1D Wasserstein distance on projected distributions.

The proposed framework for training an arbitrary image enhancement model is shown in Figure~\ref{fig:framework}. First, VGG features~\cite{simonyan2014very} for the ground truth and the predicted images are obtained. Then, the extracted feature activation maps are projected and for each projected feature map we compute the 1D Wasserstein distance between the predicted and target distributions. Finally, the sum of the 1D Wasserstein distances between the distributions is used in our learning loss. The proposed framework is straightforward, and it can be easily implemented (see Algorithm~\ref{alg:the_alg}). In comparison with the existing perceptual losses our Projected Distribution Loss (PDL) does not add any significant computational complexity.  

We demonstrate effectiveness of the proposed loss on an extensive set of experiments analyzing the effect of adopting different loss functions on five different image restorations tasks: denoising, single-image upscaling (super-resolution), demosaicing, JPEG artifact removal and deblurring. We evaluate the results according to different perceptual and distortion metrics. The proposed distribution loss leads to perceptually superior results without introducing significant distortions.

The remainder of the paper is organized as follows.
Section~\ref{sec:related_work} discusses related work and the substantial differences of what we propose. Section~\ref{sec:comparing_distributions} explains the mathematical foundations of the distribution loss, and then the details are discussed in Section~\ref{sec:distribution-loss}. Section~\ref{sec:results} presents extensive results and comparisons to other popular loss functions. Finally we conclude the paper in Section~\ref{sec:discussion}.

\section{Related work}
\label{sec:related_work}
Image restoration aims to generate a high-quality image from its degraded low-quality measurement (e.g., low-resolution, compressed, noisy, blurry).  Recently, deep convolutional neural networks (CNN) have been popular due to their remarkable performance in many image restoration tasks. A detailed analysis of the image restoration literature is beyond the scope of this article. In the following we present the most relevant work. %

\noindent \textbf{From deep features to perceptual losses.} Briefly after the resurgence of CNNs, Dong et al.~\cite{dong2015image} presented an end-to-end  convolutional neural model that super-resolves an image. The network is trained end-to-end by minimizing the mean squared error ($L_2$ loss). At that time results were remarkably good comparing to previous approaches that learn very shallow models or assumed a prior on the data (such as sparsity or small TV).  In 2016, Zhao et al.~\cite{zhao2017loss} executed a thorough experimental comparison of different pixel losses ($L_1$, $L_2$, SSIM, MS-SSIM) in three applications (super-resolution, JPEG artifacts removal, and joint demosaicing \& denoising). Their study led to proposing a loss that combines $L_1$ and MS-SSIM that obtains superior restoration results particularly in terms of image artifacts.

Johnson et al.~\cite{johnson2016perceptual} introduced the idea of using deep image features extracted from an auxiliary network to capture perceptual information. Their used deep features are the activation maps of a deep CNN trained for object recognition. Their perceptual loss directly penalizes element-wise differences in the deep feature space leading to superior perceptual results. The perceptual loss has become a de facto standard for deep supervised learning in image restoration applications. However, being also an element-wise distance also suffers from the regression to the mean phenomenon.
Zhang et al.~\cite{zhang2018unreasonable} introduced an experimental study where they showed that intermediate deep features trained for computer vision tasks capture the low-level perceptual similarity surprisingly well. An image metric is introduce to compare image pairs in a perceptual way (LPIPS). Based on similar observations, Talebi and Milanfar~\cite{talebi2018learned,talebi2018nima} proposed a trainable model to predict the distribution of human opinion scores using deep features.
A systematic analysis carried out in~\cite{tarig2020why} shows that deep features are indeed correlated with basic human perception characteristics, such as contrast sensitivity and orientation selectivity. Their findings suggest that a perceptual loss function can potentially select a subset of features that are more correlated with human perception leading to a better perception-distortion trade-off
~\cite{blau2018perception}.

Motivated by maintaining the statistics of natural images, the contextual loss (CTX) was introduced in~\cite{mechrez2018contextual}. The contextual loss compares the deep features of the generated image with the most similar ones of the reference image (similar context, does not need data to be aligned). The authors adopted this loss in the context of non-aligned image transformation and in particular style transfer. In a follow up the authors show that this loss produces high-quality results on other image processing applications such us super-resolution~\cite{mechrez2018maintaining}. The authors also show that the contextual loss can be seen as an approximation of the Kullback-Leibler divergence between the input and target features. In~\cite{zhang2019zoom} the contextual loss is modified into a contextual bilateral loss (CoBi) that prioritizes local features in the global search of similar features. \vspace{.7em}

\noindent \textbf{Adversarial losses.}
Adversarial losses~\cite{ledig2017photo,isola2017image,kupyn2018deblurgan,kupyn2019deblurgan} based on generative adversarial training~\cite{goodfellow2014generative,arjovsky2017wasserstein} have become the golden standard in terms of perceptual quality. The overall idea is to train in an alternate fashion a generator, that learns to generate realistic images and a discriminator that learns to distinguish real from fake images. The trained discriminator can be plug-in on any loss as an extra term that penalizes images that do not look realistic.  The price to pay is that in general generated images do not tightly follow the low-resolution observation leading to a superior distortion than perceptual or pixel losses~\cite{cohen2018distribution}. \vspace{.7em}%

\noindent \textbf{Our Projected Distribution Loss (PDL).}
We propose a distribution loss that penalizes difference of generated and target deep image feature distributions based on Optimal Transport theory~\cite{villani2008optimal} and in particular with the Wasserstein metric~\cite{villani2008optimal,peyre2019computational}. 
By comparing the distribution of features in addition to the pixel values, we are able to produce images with a higher level of realism. Furthermore since we compare the generated image directly with its reference, the amount of hallucinated content is better controlled than in a (non-referenced) adversarial loss.  Our loss can be seen as an intermediate alternative between the L1-perceptual loss and the adversarial losses.
Comparing high-dimensional distributions is a challenging problem. In~\cite{rabin2011wasserstein} authors introduce the sliced-Wasserstein distance as a way to circumvent the high-dimensional drawbacks, but also keep the desirable mathematical properties of the Wasserstein distance. The sliced-Wasserstein distance consists of projecting the data distribution into all possible directions and then computing the average value. 
The Wasserstein and sliced-Wasserstein have recently received attention in a number of different computer vision applications
~\cite{frogner2015learning, peyre2019computational, wu2019sliced, zhang2020deepemd}.
We use the sliced Wasserstein distance to compare the feature distributions by adding the comparison of 1D marginal distributions of projected features. The projected distribution loss works as a complement to a pixel fidelity loss term.

 \begin{figure*}[t]
    \centering
    \scriptsize

    \begin{minipage}[c]{.39\linewidth}
    \centering
    \includegraphics[width=.8\linewidth]{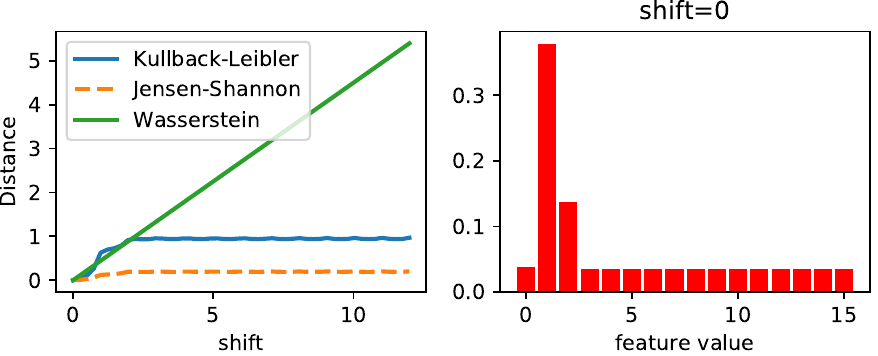}
    
    \includegraphics[width=\linewidth]{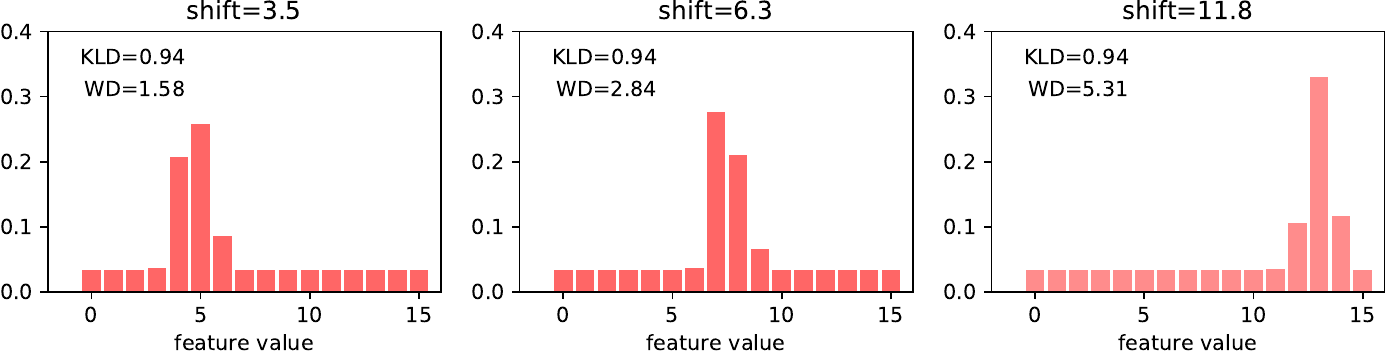}
    
    (a) Toy Example
    \end{minipage}
    \begin{minipage}[c]{.6\linewidth}
    \begin{minipage}[c]{.26\linewidth}
    \centering
    \vspace{1em}
    
    \includegraphics[clip, trim=0 10 0 10, width=.9\linewidth]{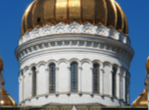}\vspace{1em}
    
    \includegraphics[clip, trim=0 40 0 40, width=.9\linewidth]{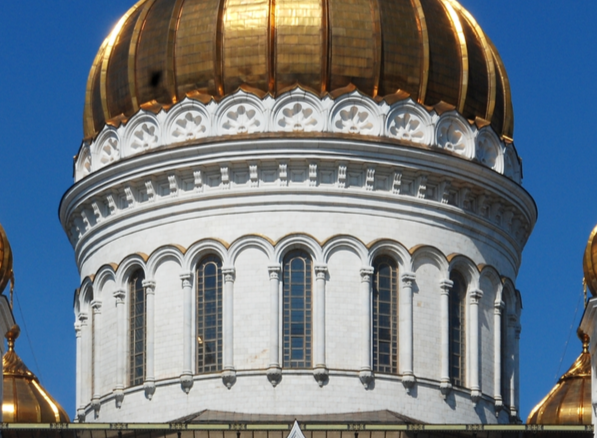}\vspace{1em} 
    
    (b) Top: LR, bottom: HR
    \end{minipage}
    \begin{minipage}[c]{.24\linewidth}
    \includegraphics[width=\linewidth]{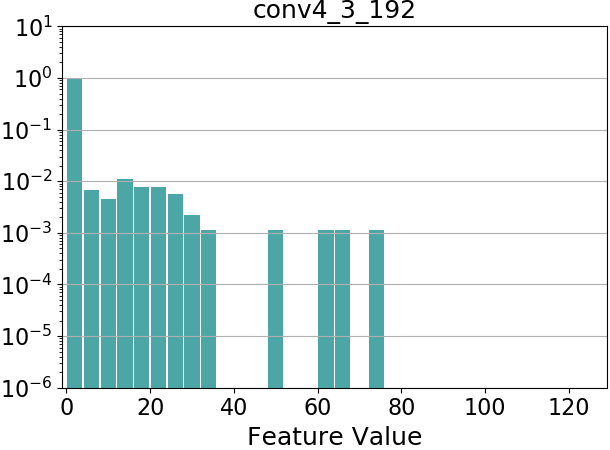} 
    \includegraphics[width=\linewidth]{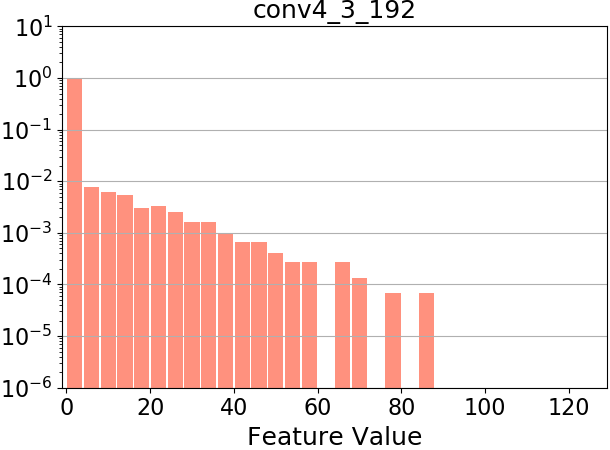} 
   
    (c) \textsc{wd}=0.10, \textsc{kld}=0.07 
    \end{minipage}
    \begin{minipage}[c]{.24\linewidth}
    \includegraphics[width=\linewidth]{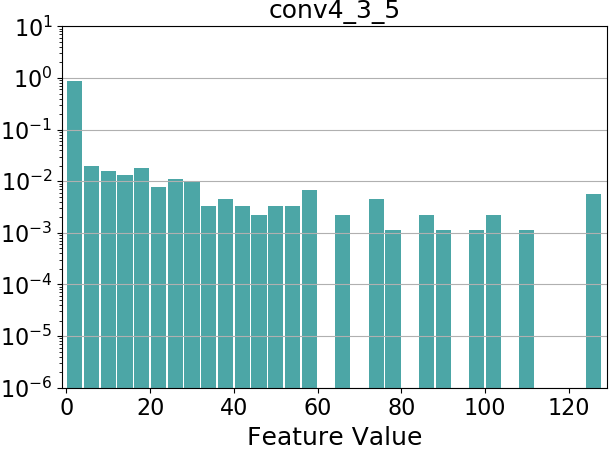} 
    \includegraphics[width=\linewidth]{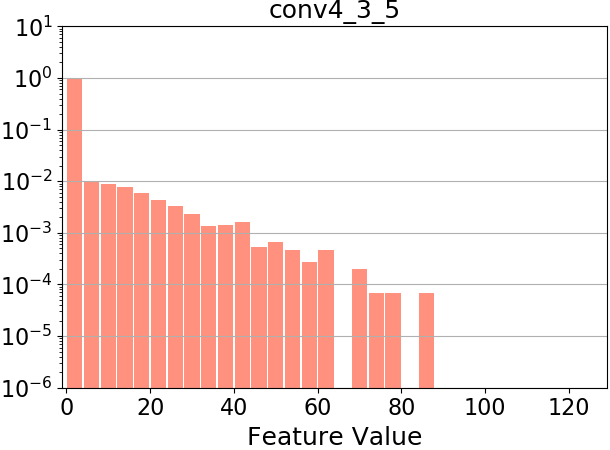} 

    (d) \textsc{wd}=0.92, \textsc{kld}=0.07 
    \end{minipage}
    \begin{minipage}[c]{.24\linewidth}
    \includegraphics[width=\linewidth]{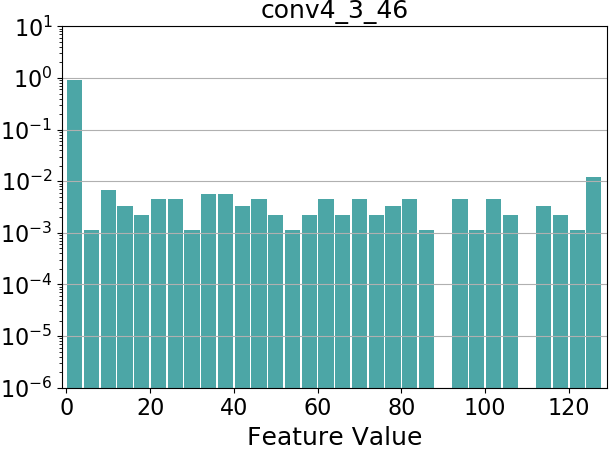} 
    \includegraphics[width=\linewidth]{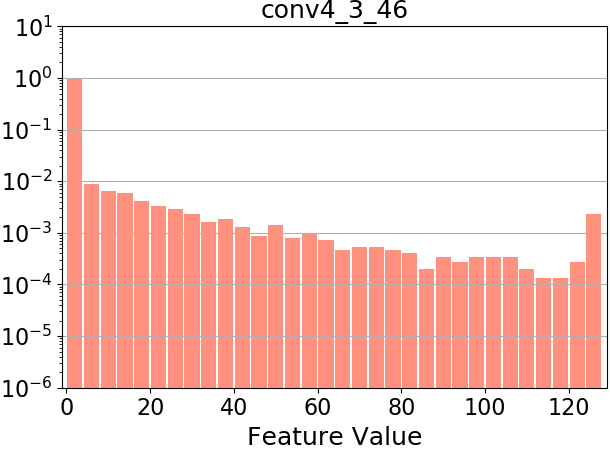} 

    (e) \textsc{wd}=1.21, \textsc{kld}=0.06
    \end{minipage}

    \end{minipage}
    \vspace{.5em}
    
    \caption{Distance between distributions. (a) toy example showing how the Wasserstein distance is sensitive to the geometry of the space. A base distribution is translated, while the KLD and JSD divergences remain constant, the Wasserstein distance increases with the shift amount. Since distant features are in fact less similar the Wasserstein distance is a natural way to compare feature distributions. (b-e) Examples of distances between distribution of VGG16 features. The first row shows some histograms from a low resolution input, and the second row shows the histogram counterparts computed at $4\times$ higher resolution. These results show how the Wasserstein distance (WD) is more sensitive to the shape of the feature distributions.}
    \label{fig:histogram_distances}
\end{figure*}

\section{Comparing feature distributions}
\label{sec:comparing_distributions}

There are several ways to define a distance (or a quasi-distance) between distributions~\cite{deza2006dictionary}. Among the most popular ones are the Kullback-Leibler (KLD) and Jensen-Shannon (JSD) divergences, the total variation, and the Wasserstein distance. 
A relevant characteristic of the Wasserstein distance is that it uses the geometry of the domain  to explicitly penalize the differences between the distributions. In Figure~\ref{fig:histogram_distances} (a) we present a toy example where a base discrete distribution (histogram) is shifted a variable amount. As the distribution shifts, the KLD and the JSD divergences remain constant while the Wasserstein distance increases. Examples of distances between distributions of VGG16 features~\cite{simonyan2014very} are shown in Figure~\ref{fig:histogram_distances} (b)-(e). The first row shows some distributions from a low resolution input, and the second row shows the high-resolution counterparts. These results show how the Wasserstein distance (measured as the Earth Mover's Distance) is more sensitive to the shape of the distributions.

Since our goal is to compare distributions of features, and given that distributions by nature belong to a metric space, we adopt the Wasserstein distance. In the following we present a short summary explaining the Wasserstein distance and the efficient variant that leads to our projected distribution loss.

\subsection{Optimal Transport and the Wasserstein distance}
Our proposed loss can be formulated as a problem of optimal transport in a metric space. 
Let $I_u$ and $I_v$ be two probability density functions (pdf) defined on $\mathbb{R}^d$. This two pdf represent the two distributions that we want to compare. The goal of optimal transport is to find a coupling $\pi$ (also known as a transport plan) that transforms $I_u$ into $I_v$ with minimal cost. The solution to this problem leads to the $p$-Wasserstein distance between $I_u$ and $I_v$,
\begin{align}
\label{eq:wasserstein}
W^p_p(I_u,I_v) =  \inf_{\pi \in \Pi(I_u,I_v)} \int_{\mathbb{R}^d \times \mathbb{R}^d} \| x - y\|^p \pi(x,y) dxdy, 
\end{align}
where $\Pi(I_u,I_v)$ denotes the space of all joint distributions $\pi$ having marginals $I_u$ and $I_v$. Unfortunately, the optimal transport problem in $\mathbb{R}^d$ does not have a closed form solution and an optimization scheme is needed~\cite{peyre2019computational}. \vspace{.7em}

\subsection{The 1D Wasserstein distance} 
In the particular case where the densities are defined on the real line, the one-dimensional p-Wasserstein distance has a closed form. In this case, it can be shown that,
\begin{align}
\label{eq:onedim_wasserstein}
W^p_p(I_u,I_v) =  \int_0^1 \left| F^{-1}_u(s) - F^{-1}_v (s) \right|^p  ds ,
\end{align}
where $F_u(s), F_v(s)$ are the cumulative distribution functions (CDF) of $I_u$ and $I_v$, respectively~\cite{peyre2019computational}. %
\vspace{.7em}

\noindent \textbf{Numerical Implementation.} In our setting, we want to compare the empirical distributions of the projected features from two images. Let $\{a_i\}_{i=1}^n$ and $\{b_i\}_{i=1}^n$ represent the set of one-dimensional (projected) features of the two images to compare. In this case, the cumulative distribution functions are directly replaced by the empirical distributions, that is,
$F_\mathbf{a}(s) = \frac{1}{n}\sum_{i=1}^n \mathbf{1}_{\{a_i \le s\}}$ and $F_\mathbf{b}(s) = \frac{1}{n} \sum_{i=1}^n \mathbf{1}_{\{ b_i \le s\}}$,
where $\mathbf{1}_{\mathcal{X}}$ is the indicator function of set $\mathcal{X} \subset \mathbb{R}$. Let $(a_{i_p})$ be the set $\{a_i\}$  sorted in ascending order, i.e., $a_{i_p} \le a_{i_{p+1}}$ (and similarly $(b_{i_p})$). The empirical distribution $F_\av(s)$ is a non-negative non-decreasing step-wise function having $n$ steps at $s_i=a_{i_p}$ and values $F(s_i) = i/n$. Then~\eqref{eq:onedim_wasserstein} can be directly computed by measuring the distance between the sorted sequences, 
\begin{align}
W^p_p(\mathbf{a},\mathbf{b}) =  \int_0^1 \left| F^{-1}_\textbf{a}(s) - F^{-1}_\textbf{b} (s) \right|^p ds  = \sum_{p=1}^n  |a_{i_p}-a_{i_p}|^p. 
\label{eq:onedim_wasserstein_sort}
\end{align}
This implies that for the one-dimensional case, we can compute the Wasserstein distance between the cumulative distributions by sorting the (projected) features. This leads to a straightforward implementation of the 1D-Wasserstein distance in Tensorflow with just a few lines of code. 
Note that sorting is not a fully-differentiable operation. In fact, $\text{sort}(\textbf{u}) = \textbf{P} \textbf{u}$, where $\textbf{P}$ is a permutation matrix that depends on $\textbf{u}$. During the forward pass the right permutation matrix $\textbf{P}$ is computed. During the backward pass, $\textbf{P}$ is kept constant, leading to an approximation of the derivative\footnote{This is straightforward to implement using the Tensorflow function \texttt{tf.nn.top\_k($\ldots$, sorted=True)}.}. 

\begin{algorithm}[t]
    \SetKwInOut{Input}{Input}
    \SetKwInOut{Output}{Output}

    \Input{Predicted image $\uv \in \mathbb{R}^{n}$, ground truth $\vv \in \mathbb{R}^{n}$, projection matrix $\textbf{W} \in \mathbb{R}^{m' \times m}$, the Wasserstein weight $\lambda$}
    \Output{Learning loss $L_\textsc{pdl}(\uv, \vv)$}
    Compute the VGG features: $\Phi(\uv)$ and $\Phi(\vv) \in \mathbb{R}^{n \times m}$\\
    Projection: $\Phi'(\uv) = \Phi(\uv) \textbf{W}^T$ and $\Phi'(\vv) = \Phi(\vv) \textbf{W}^T$\\
    Pixel fidelity term: $L_\textsc{pdl}(\uv, \vv) = \|\uv - \vv\|_q$\\
    \For{$j = 1$ to $m'$} {
         (1D Wasserstein)\\
         Sort $\phi'_j(\uv)$ and $\phi'_j(\vv)$\\
         $L_\textsc{pdl}(\uv, \vv) \:+= \lambda \|\phi'_j(\uv) - \phi'_j(\vv)\|_p$
    }
    \caption{The proposed PDL training loss.}
    \label{alg:the_alg}
\end{algorithm}

\begin{figure*}[t]
\vspace{-0 mm}
\begin{center}
\includegraphics*[viewport=1 170 720 420,scale=0.7]{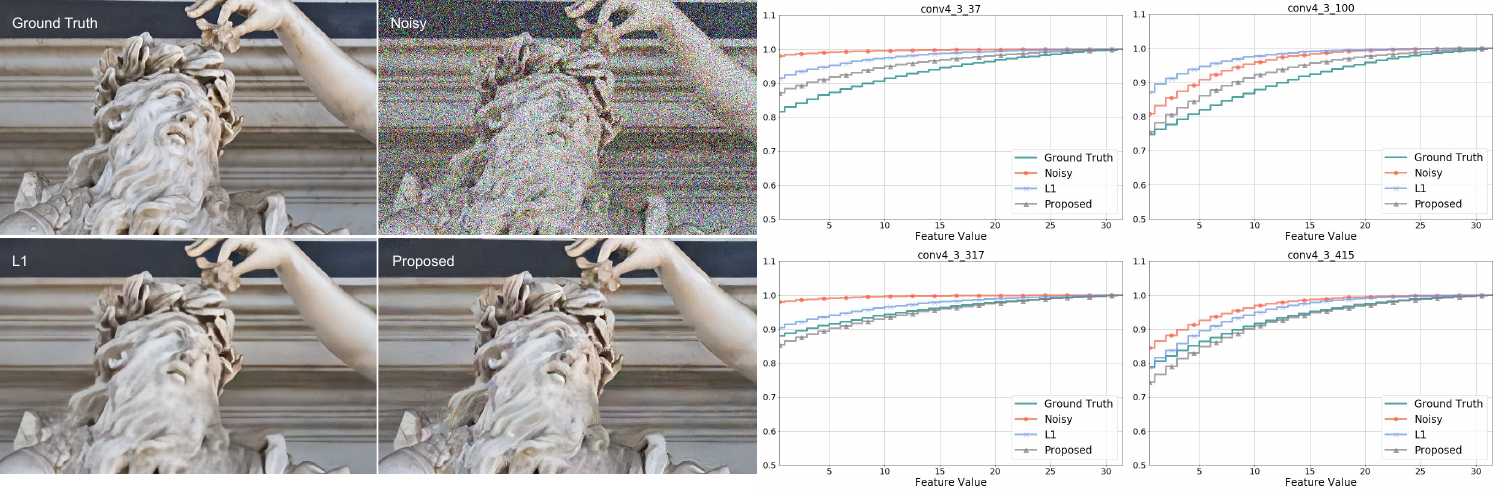}
\end{center}
\vspace{-2 mm}
{\caption{The cumulative distribution of VGG features. The enhancement CNNs are trained with two different losses: 1) L1 norm of the VGG feature differences (labeled as L1), 2) the sliced-Wasserstein distance of the extracted VGG features (labeled as proposed). As can be seen in the plots, compared to the L1 case, the cumulative distribution of the features obtained from our proposed method is closer to the ground truth. This leads to a perceptually superior result with more details.} \label{fig:cum_hists}}
\vspace{-4 mm}
\end{figure*}

\vspace{.7em}

\subsection{Sliced Wasserstein distance and generalizations}
Bonneel et al.~\cite{bonneel2015sliced} proposed to exploit the goodness of the  one-dimensional case and introduced the sliced-Wasserstein distance. By making use of the Radon transform, they show that one can integrate one-dimensional Wasserstein distances and define a distance between multi-dimensional distributions. Let $\theta \in \mathbb{S}^{d-1}$, we define the $\theta$-projector operator as $p_\theta(x) = \theta \cdot x$. 
This leads to the marginal distribution $p^\ast_\theta I_u(s) = \int_{\mathbb{R}^d} I_u(x)\delta(s - p_\theta(x))dx$.
Then, given $I_u,I_v$ defined in $\mathbb{R}^d$, the sliced-Wasserstein distance is defined as
\begin{align}
\label{eq:sliced_wasserstein}
\mbox{SW}^p_p(I_u, I_v) = \int_{\mathbb{S}^{d-1}} W_p \left( p^\ast_\theta I_u, p^\ast_\theta I_v \right) d \theta.
\end{align}
In practice, the integral in~\eqref{eq:sliced_wasserstein} is approximated by Monte Carlo sampling and averaging the 1D Wasserstein distance on a set of random projections.

The sliced distance has recently received significant attention~\cite{wu2019sliced} since it inherits all the desirable properties of the multidimensional Wasserstein distance but it presents an alternative and more efficient calculation. The main drawback is that the number of necessary samples to accurately approximate the integral grows exponentially with respect to the dimension. In particular most directions will not present relevant information for discriminating the two distributions~\cite{kolouri2019generalized}. Several works propose to mitigate this problem by finding discriminative directions. In~\cite{deshpande2018generative} authors propose to use a discriminator, similar as used in GANs, to provide good discriminant projections. \cite{deshpande2019max} propose an alternative to the sliced Wasserstein distance that replace the average over the set of random projections with the maximum value. \cite{kolouri2019generalized} introduce a generalization of the sliced Wasserstein distance using non-linear projections.

Our proposed Projected Distribution Loss (PDL) is based on the 1D Wasserstein distance for comparing the VGG feature distributions of the generated and target image on a set of one-dimensional projections of the extracted feature maps.

\section{Projected Distribution Loss} 
\label{sec:distribution-loss}

The proposed learning loss is summarized in Algorithm~\ref{alg:the_alg}. Starting with the predicted image $\uv$, the extracted  features (e.g., VGG16-convN activation map~\cite{simonyan2014very}) are represented by $\Phi(\uv) \in \mathbb{R}^{n \times m}$, where $n = h\times w$ represent the spatial dimensions, and $m$ is the number of extracted features.  Similarly, the features associated with the target image $\vv$ are denoted by $\Phi(\vv)$.

\noindent \textbf{Feature projection.} We are aiming to compare the feature distributions, thus, we assume that features of image $\uv$ are presented as set of $n$ vectors $\phiv_i(\uv) \in  \mathbb{R}^m$ with $i=1,\ldots,n$.  To compute the distance between multidimensional distributions we can aggregate one-dimensional distances computed by projecting the features into different directions.  The sliced-Wasserstein distance is computed by projecting the features into random directions on the sphere. Let $\wv_j \in \mathbb{R}^m$ such that $\|\wv_j\|$ = 1 with $j=1,\ldots, m'$ directions in $\mathbb{R}^m$. Then, the features are projected to generate $\phi'_{i,j} (\uv) = \wv_j^T  \phiv_i(\uv)$. An alternative naive procedure is to compute the distances in the distributions taken from each individual feature value. This implies computing the distance between the marginal distributions. This makes sense if the features are independent.

In this work we follow the naive approach of computing and averaging the distance between the marginal distributions of the features.  Nonetheless, in the experimental section we present an analysis comparing different (random) projections schemes.

\subsection{Projected Distribution Training Loss}
Given an image $\uv$, its target counterpart $\vv$, and a set of $m'$ projections $\{\wv_j\}$, our training loss is defined as
\begin{align}
\label{eq:pdl-loss}
L_\textsc{pdl}(\uv, \vv) = \|\uv - \vv\|_q + \lambda \sum_{j=1}^{m'} W_p \left(\phiv'_j(\uv), \phiv'_j(\vv) \right),
\end{align}
where we have compactly denoted by $\phiv'_j = \{\phi'_{1,j},\ldots, \phi'_{n,j}\}$ to the set of projected features at direction $\wv_j$ for the image $\uv$ and $\vv$, respectively. The distance $W_p$ is computed using~\eqref{eq:onedim_wasserstein_sort} and the non-negative constant $\lambda$ controls the effect of the distribution loss. In all the presented results we used $q=1$ and $p=1$. Note that this loss is fully differentiable and straightforward to implement. The impact of the regularization parameter in (\ref{eq:pdl-loss}) is discussed in the next section. 
The proposed distribution loss combines a pixel fidelity loss term with the distribution mismatch penalization that allows to transfer details (texture, grain) without forcing them to be located in the exact same spatial position as in the target reference. This flexibility allows mitigating the \emph{regression to the mean} problem, where, for example, slightly different high-quality images with identical feature distributions lead to very similar low resolution images. In this particular case, any point loss (pixel or $L_1$ on features) as the perceptual loss will learn to predict the average leading to a less realistic (blurry) image. Figure~\ref{fig:cum_hists} shows examples comparing the proposed loss with the original perceptual loss $L_\text{percep}$. Penalizing the distance between distributions results in better preservation of fine details and overall sharpness; whereas using the $L_1$ norm on the features leads to over-smoothed images.

\vspace{.3em}

\noindent \textbf{Perceptual Loss}. The (original) perceptual loss~\cite{johnson2016perceptual} is computed by replacing the distribution term in~\ref{eq:pdl-loss} by the $L_p$ distance directly computed on the extracted features,
\begin{align}
\label{eq:percep-loss}
L_\text{percep}(\uv, \vv) = \|\uv - \vv\|_q + \lambda \| \Phi(u) - \Phi(v)\|_p.
\end{align}

As we will show in the experimental part, measuring the distance directly between the features produces artifacts in particular in problems that are seriously ill-conditioned (such as denoising under strong noise).

\vspace{.3em}

\noindent \textbf{Contextual Loss}
Our proposed framework can be closely compared with the contextual loss \cite{mechrez2018contextual}. This loss is based on the idea that for calculating similarity between two images one should find corresponding features with minimal distances from each other. To this end, the authors define the contextual similarity based on the maximum dot product (CTXDP), or alternatively the minimum $L_2$ distance (CTXL2) between features. Also, it has been shown that the contextual loss is an approximation of the KL divergence~\cite{mechrez2018maintaining}. Since our framework is based on the Wasserstein distance between image features, in the following section we carry out detailed comparisons with the contextual loss. We show that in most image enhancement scenarios, the proposed Wasserstein loss shows a consistent perceptual advantage over the contextual loss.

\section{Experiments}
\label{sec:results}
In what follows we present several experiments comparing the proposed PDL distribution loss against other perceptual losses. As a baseline we train a model per application without any perceptual loss and just an $L_1$ fit to the pixel values ($\lambda=0$ in~\eqref{eq:pdl-loss}). We also compare to the $L_1$-perceptual loss given by~\eqref{eq:percep-loss} and the recently introduced contextual loss~\cite{mechrez2018contextual}. In all experiments, we tried to find the best balance between the data fitting term and the perceptual term that produces the best possible results, which is a complex and expensive task. 

Since we focus on comparing different training loses, all models are trained using the same model architecture and optimization parameters. All perceptual loses are computed on \texttt{VGG16\-conv4} features. All models, unless otherwise stated, are trained for $n_\text{iter}=10^6$ iterations, using ADAM optimizer with default parameters and a mini-batch of size $8$.
 
We extend our analysis to five image enhancement tasks: image denoising, single image super-resolution, deblurring, JPEG artifacts removal and demosaicing.  All the experiments are carried out using the standard Div2k dataset~\cite{agustsson2017ntire}. In all applications we simulated the respective degradation operator to generate training and validation data. The evaluation is done in the DIV2K validation dataset. For each application we generated $200,000$ random $256\times 256$ crops using the div2k training images. This implies that the feature distributions are locally compared to the given image crop.

\subsection{CNN Models}

We focus on five image enhancement applications, and use three CNN models to showcase performance of the proposed learning loss. For denoising, JPEG artifact removal and demosaicing we use the SRN~\cite{tao2018scale} model, which consists of typical convolutional and residual blocks~\cite{he2016deep,nah2017deep} at multiple spatial scales (i.e., encoder-decoder network~\cite{ronneberger2015u,mao2016image,tao2018scale}). 

We chose this model since it proves to work well for high noise levels. For the super-resolution application we use the EDSR~\cite{lim2017enhanced}, which applies depth-to-space operation to increase the spatial resolution of the convolutional feature maps. Also, for deblurring the DsDeblur~\cite{gao2019dynamic} model is employed in our experimentation. This model is a 3-stage encoder-decoder architecture with a selective parameter sharing scheme. In our experimentation we opt to use the default architectures proposed by respective authors.

\noindent \textbf{(R3,R4) No-reference measures (NIQE, FID).} We have added FID and NIQE scoresto the main tables in the paper to provide the reader with a better sense of the perceptual quality of the reconstructed images. NIQE is a no-reference quality metric, and FID is a no-reference distributional comparison metric popular in generative modeling. The updated results are given in

\subsection{Quantitative and Qualitative Evaluation}
To evaluate our results we report three full-reference metrics, that is, PSNR, MS-SSIM~\cite{wang2003multiscale}, and LPIPS~\cite{zhang2018unreasonable}, as well as the no-reference image quality scores NIQE, and the Fr\'echet Inception Distance (FID)\footnote{FID is computed using \href{http://github.com/mseitzer/pytorch-fid}{pytorch-fid}, and NIQE using \href{www.mathworks.com/help/images/ref/niqe.html}{Matlab-niqe}.}, a no-reference distributional comparison metric.
Note that in contrast to PSNR and MS-SSIM, lower LPIPS means higher similarity to the ground truth. Similarly, a lower NIQE and FID scores implies a better restoration. We should mention that this particular way of weighting the different metrics may seem somewhat arbitrary, however, we found that it sufficiently serves the purpose of finding the best model configuration.\vspace{.3em}

\noindent \textbf{User study.} In addition to the quantitative metrics mentioned above, we also run perceptual studies with human subjects. 
We used the forced-choice pairwise comparison framework through Amazon Mechanical Turk, and assigned our tasks to 25 human raters with a minimum of 70\% approval rating. These raters were paid 2 cents per question, and were asked to select the image with the best quality from side-by-side image crops of size $800\times800$. We also asked raters to only use displays with resolution $1920\times1080$ or higher for our experiment. Each rater answered 20 questions to compare results from two models. To avoid potential rater bias in ratings, images in each pair were randomly permuted and displayed. The reported results in the paper represent the average raters preference computed from 500 comparisons (20 raters, and 25 pairs). Since the outcome of each pairwise comparison can be interpreted as a Bernoulli random variable with probability $p$, the standard deviation can be expressed as $\sqrt{p(1-p)}$. This means that the standard error for each reported average is $\sqrt{\nicefrac{p(1-p)}{n}}$ with $n=500$ representing the total number of trials. Note that in the worst case scenario where $p=0.5$, the standard error is about $0.022$.\vspace{.3em}

\subsection{Computational cost of the PDL loss} 
VGG16 is a heavy network that introduces an additional cost compared to the classical pixel loss. In principle, the additional cost of our PDL loss over the perceptual loss can be attributed to the sorting operation ($O(N \log N)$). Table~\ref{tab:training-time} shows the relative training speeds in global (gradient descent) steps per second for each of the tested losses. Interestingly, in practice PDL and the classical perceptual $L_1$ loss have very similar training times which indicates that the cost of sorting is marginal. Additionally, all the VGG-based perceptual losses run approximately 50\% slower than the baseline. It is worth noting that all this extra cost is only applied at training time, and the inference complexity is the same across all loss options explored for each application.
\begin{table}[h!]
\begin{center}
\footnotesize
\renewcommand{\arraystretch}{0.7}
\setlength{\tabcolsep}{3.5pt}
\begin{tabular}{rcccccc} 
             & No-perceptual &	L1 & L2 & CTXDP & CTXL2 & PDL \\\cmidrule{1-7}
steps/sec    & 1.00	& 0.57 & 0.54 &	0.54 & 0.53 & 0.57 
\end{tabular}
\end{center}
\caption{Relative training speed in global\_step/sec.} \vspace{-2.5em}
\label{tab:training-time}
\end{table}

\subsection{Denoising under strong noise}
To analyze the proposed loss we focus on the fundamental problem of image denoising. We address the particular and challenging case where the input image has been severely damaged by additive Gaussian noise (noise std. $\sigma=100$). This allows us to get a clearer idea of the differences of the compared methods.

\begin{figure*}
    \centering
    \scriptsize
    
    \begin{minipage}[c]{\linewidth}
    \begin{tikzpicture}
      \node[anchor=north west,inner sep=0] (image) at (0,0) {\includegraphics[clip, trim=543 430 150 160, width=.252\linewidth]{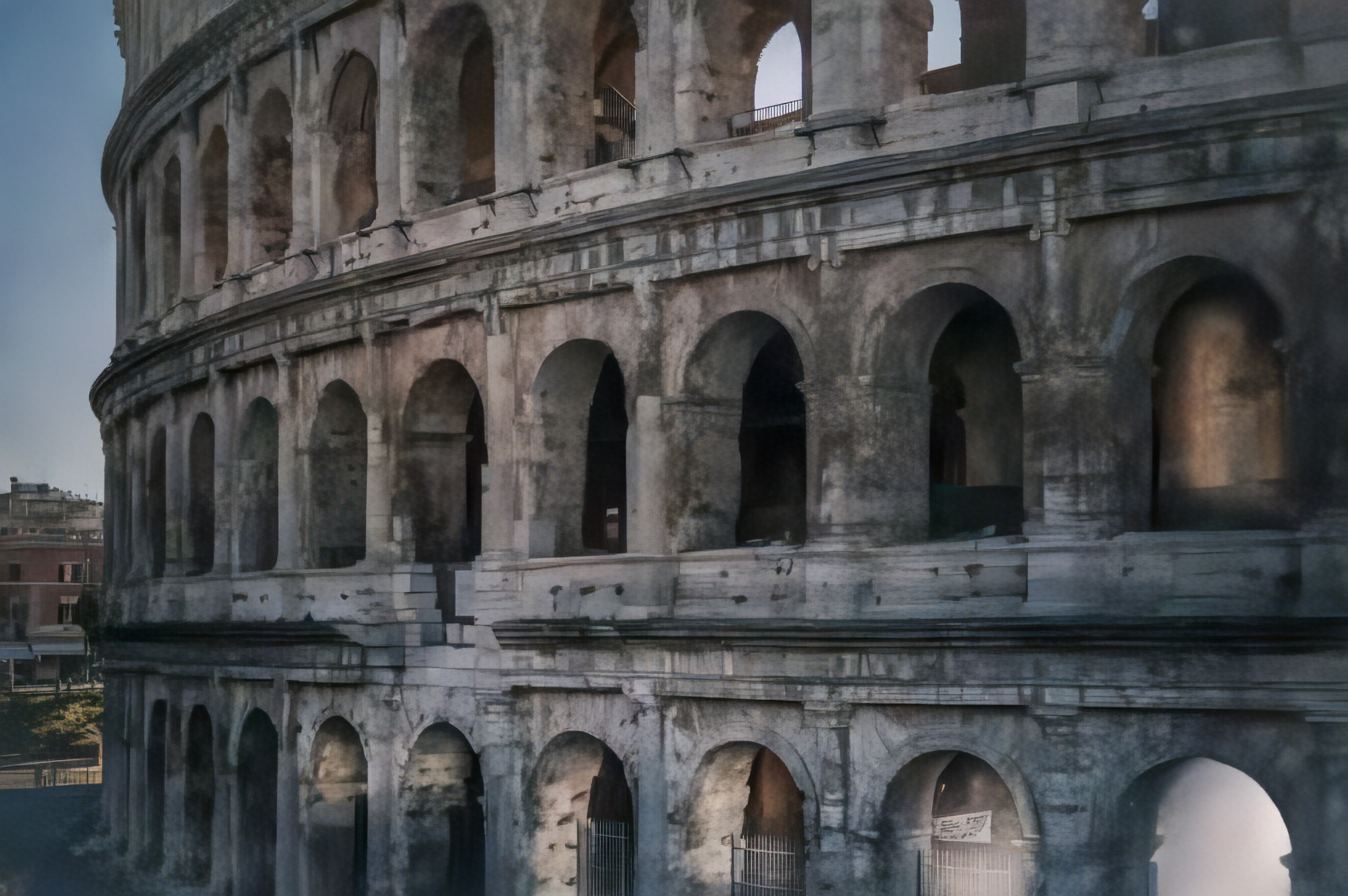} };
      \begin{scope}[x={(image.north east)},y={(image.south west)}]
        \draw[yellow1, very thick](0.3319, 0.2469) rectangle (0.4708, 0.4950);
        \node[] at (0.12,0.935) {\begin{color}{white}L1=0.01\end{color}};
      \end{scope}
    \end{tikzpicture} 
    \begin{overpic}[width=.145\linewidth]{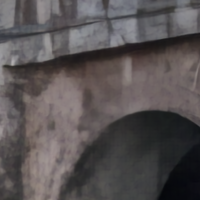}
    \put(2,2){\begin{color}{white}L1=0.001\end{color}}
    \end{overpic}
    \begin{overpic}[width=.145\linewidth]{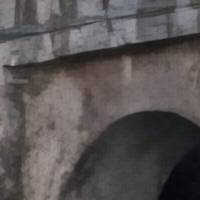}
    \put(2,2){\begin{color}{white}L1=0.005\end{color}}
    \end{overpic}
    \begin{overpic}[width=.145\linewidth]{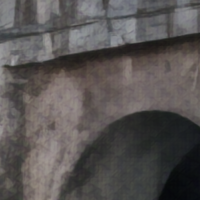}
    \put(2,2){\begin{color}{white}L1=0.01\end{color}}
    \end{overpic}
    \begin{overpic}[width=.145\linewidth]{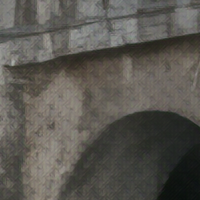}
    \put(2,2){\begin{color}{white}L1=0.1\end{color}}
    \end{overpic}
    \begin{overpic}[width=.145\linewidth]{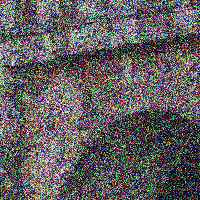}
    \put(2,2){\begin{color}{white}noisy\end{color}}
    \end{overpic}
    \end{minipage} 

\vspace{.15em}    

    \begin{minipage}[c]{\linewidth}
    \begin{tikzpicture}
      \node[anchor=north west,inner sep=0] (image) at (0,0) {\includegraphics[clip, trim=543 430 150 160, width=.252\linewidth]{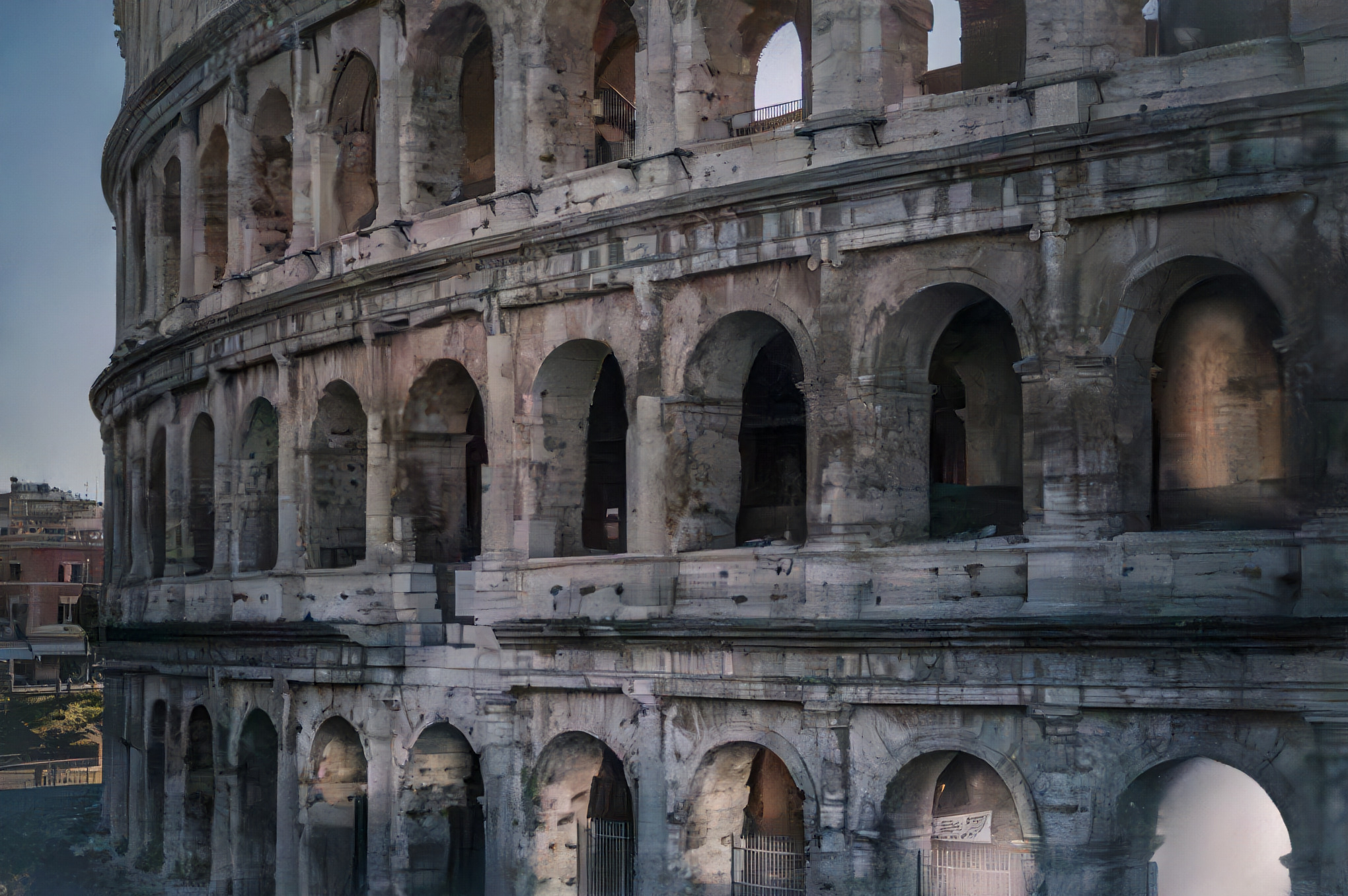} };
      \begin{scope}[x={(image.north east)},y={(image.south west)}]
        \draw[yellow1, very thick] (0.3319, 0.2469) rectangle (0.4708, 0.4950);
        \node[] at (0.15,0.935) {\begin{color}{white}PDL=0.01\end{color}};
      \end{scope}
    \end{tikzpicture} 
    \begin{overpic}[width=.145\linewidth]{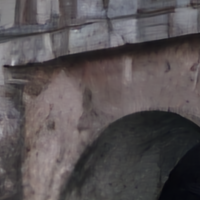}
    \put(2,2){\begin{color}{white}PDL=0.001\end{color}}
    \end{overpic}
    \begin{overpic}[width=.145\linewidth]{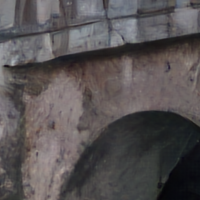}
    \put(2,2){\begin{color}{white}PDL=0.005\end{color}}
    \end{overpic}
    \begin{overpic}[width=.145\linewidth]{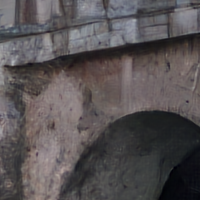}
    \put(2,2){\begin{color}{white}PDL=0.01\end{color}}
    \end{overpic}
    \begin{overpic}[width=.145\linewidth]{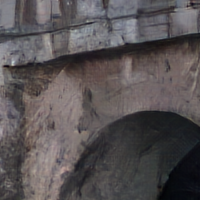}
    \put(2,2){\begin{color}{white}PDL=0.02\end{color}}
    \end{overpic}
    \begin{overpic}[width=.145\linewidth]{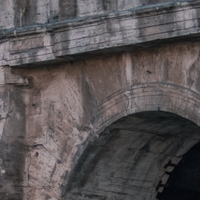}
    \put(2,2){\begin{color}{white}reference\end{color}}
    \end{overpic}
    \end{minipage}

\vspace{.3em}
      \begin{minipage}[c]{\linewidth}
    \begin{tikzpicture}
      \node[anchor=north west,inner sep=0] (image) at (0,0) {\includegraphics[clip, trim=100 330 650 280, width=.252\linewidth]{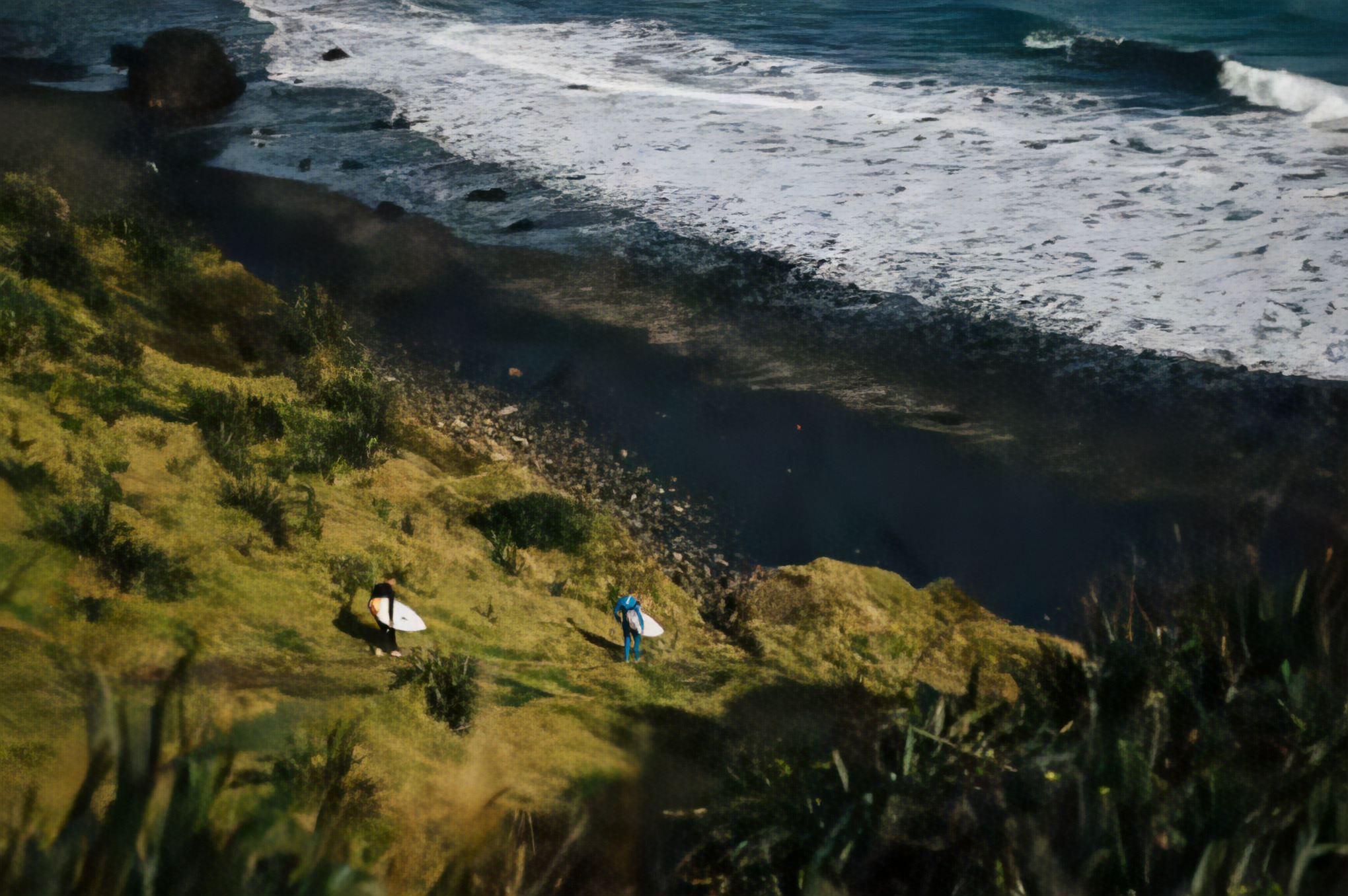} };
      \begin{scope}[x={(image.north east)},y={(image.south west)}]
        \draw[red1, very thick](0.2380, 0.3834) rectangle (0.3930, 0.6515);
        \node[] at (0.12,0.935) {\begin{color}{white}L1=0.01\end{color}};
      \end{scope}
    \end{tikzpicture} 
    \begin{overpic}[width=.145\linewidth]{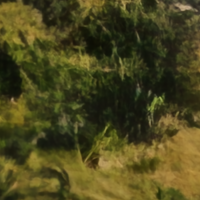}
    \put(2,2){\begin{color}{white}L1=0.001\end{color}}
    \end{overpic}
    \begin{overpic}[width=.145\linewidth]{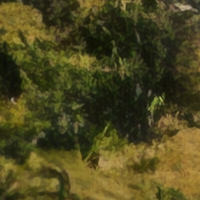}
    \put(2,2){\begin{color}{white}L1=0.005\end{color}}
    \end{overpic}
    \begin{overpic}[width=.145\linewidth]{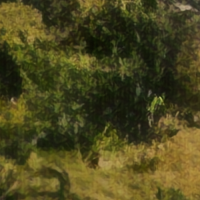}
    \put(2,2){\begin{color}{white}L1=0.01\end{color}}
    \end{overpic}
    \begin{overpic}[width=.145\linewidth]{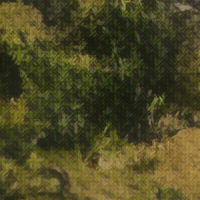}
    \put(2,2){\begin{color}{white}L1=0.1\end{color}}
    \end{overpic}
    \begin{overpic}[width=.145\linewidth]{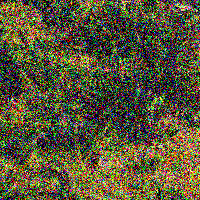}
    \put(2,2){\begin{color}{white}noisy\end{color}}
    \end{overpic}
    \end{minipage} 

\vspace{.15em}    

    \begin{minipage}[c]{\linewidth}
    \begin{tikzpicture}
      \node[anchor=north west,inner sep=0] (image) at (0,0) {\includegraphics[clip, trim=100 330 650 280, width=.252\linewidth]{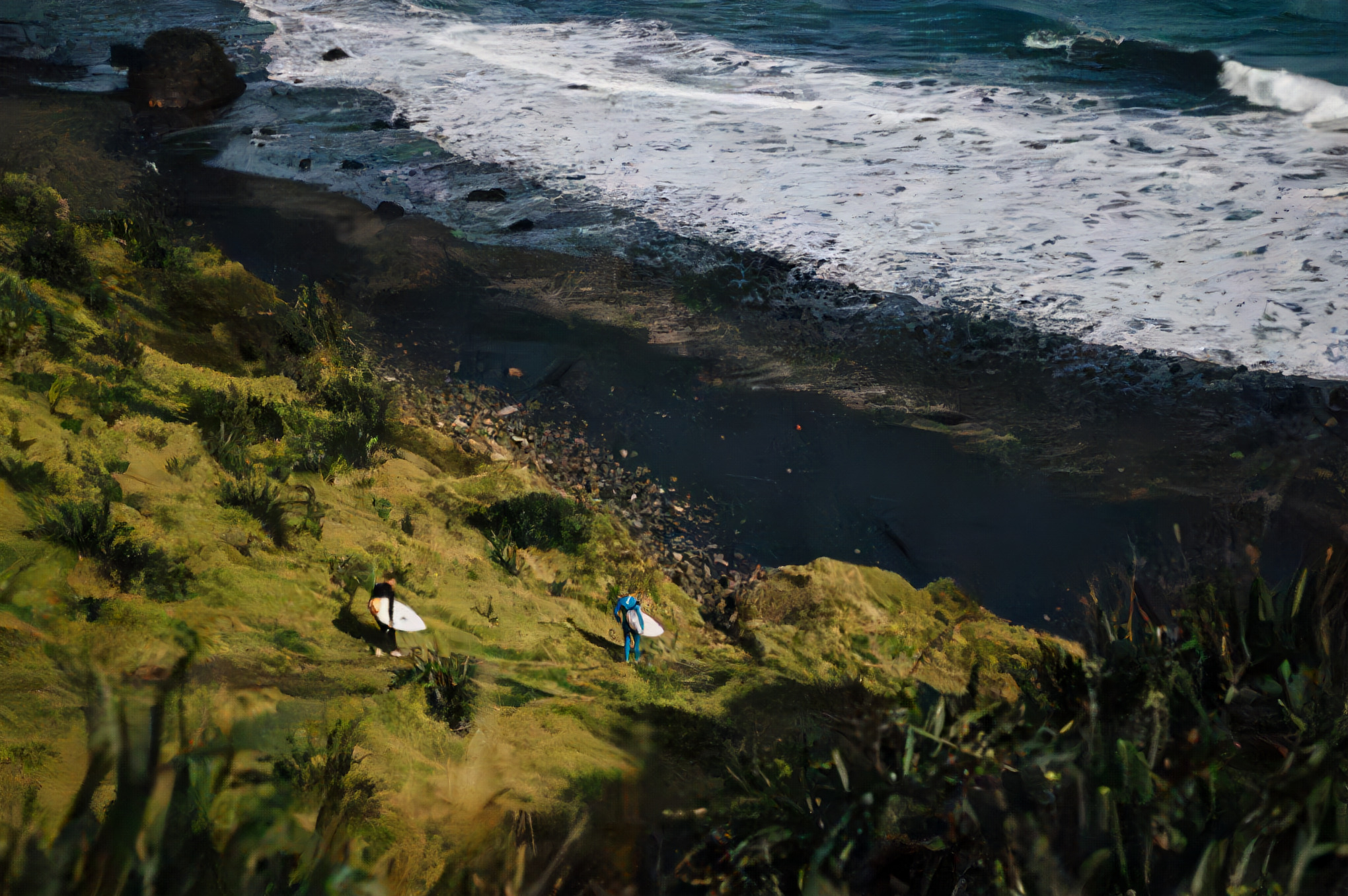} };
      \begin{scope}[x={(image.north east)},y={(image.south west)}]
        \draw[red1, very thick] (0.2380, 0.3834) rectangle (0.3930, 0.6515);
        \node[] at (0.15,0.935) {\begin{color}{white}PDL=0.01\end{color}};
      \end{scope}
    \end{tikzpicture} 
    \begin{overpic}[width=.145\linewidth]{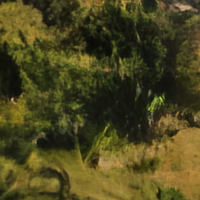}
    \put(2,2){\begin{color}{white}PDL=0.001\end{color}}
    \end{overpic}
    \begin{overpic}[width=.145\linewidth]{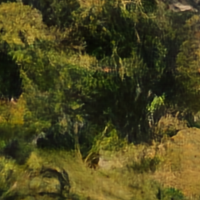}
    \put(2,2){\begin{color}{white}PDL=0.005\end{color}}
    \end{overpic}
    \begin{overpic}[width=.145\linewidth]{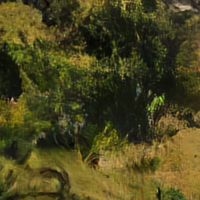}
    \put(2,2){\begin{color}{white}PDL=0.01\end{color}}
    \end{overpic}
    \begin{overpic}[width=.145\linewidth]{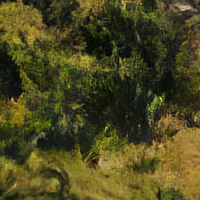}
    \put(2,2){\begin{color}{white}PDL=0.02\end{color}}
    \end{overpic}
    \begin{overpic}[width=.145\linewidth]{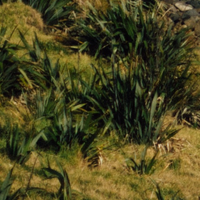}
    \put(2,2){\begin{color}{white}reference\end{color}}
    \end{overpic}
    \end{minipage}

    \caption{Denoising under strong noise. Comparison of PDL and L1-perceptual loss. For each shown result we indicate the weight value for the $L_1$-perceptual/PDL loss term.}
    \label{fig:l1-vs-pdl}
\end{figure*}
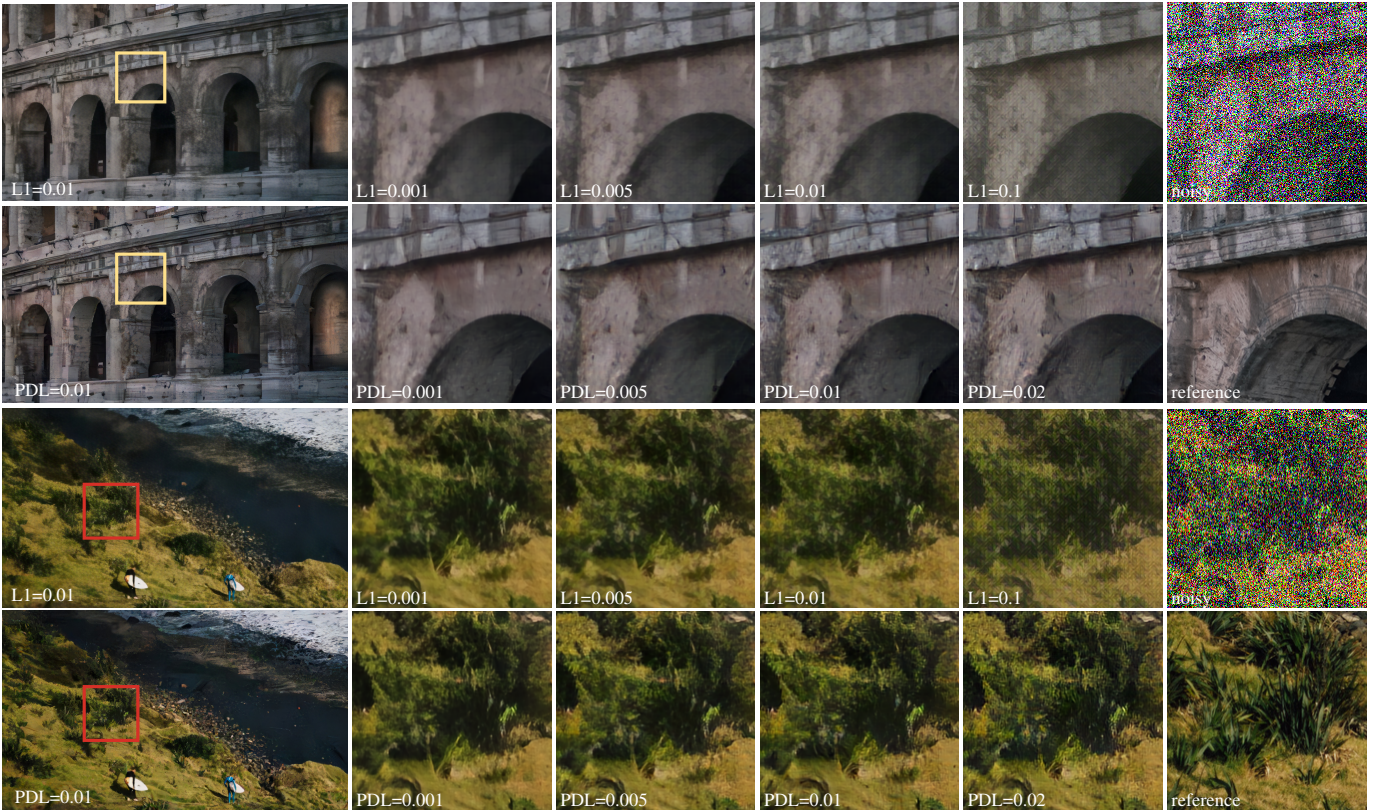

\vspace{.3em}
\noindent \textbf{Projected Distribution Loss vs $L_1$-perceptual loss~\cite{johnson2016perceptual}.} Our first and base experiment is to compare the proposed PDL distribution loss against the classical perceptual loss~\cite{johnson2016perceptual}. This is a key experiment showing the importance of comparing feature distributions over comparing extracted (aligned) features directly. 
Comparing training losses with multiple terms is challenging as it requires choosing a relative weight between each loss term. To accomplish this we train several models using different weights for each loss (eqs.~\eqref{eq:pdl-loss} and \eqref{eq:percep-loss}). In Table~\ref{tab:denoising-s100} we present a summary of the average quantitative performance on the validation set. As can be seen, the proposed distribution loss produces significantly better LPIPS values with similar PSNR values. The weight parameter in each loss plays a major role between the pixel distortion and the perceived quality measured by LPIPS. %
In Figure~\ref{fig:l1-vs-pdl} we show some results for the different trained models. Our proposed PDL produces images with more fine texture and overall sharpness. Increasing the contribution of the L1 perceptual term in~\eqref{eq:percep-loss} does not improve the overall results but leads to images with unrealistic artifacts. 

\begin{figure*}
    \centering
    \scriptsize
    \begin{overpic}[width=.162\linewidth]{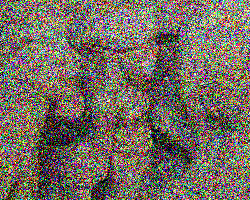}
    \put(2,2){\begin{color}{white}noisy\end{color}}
    \end{overpic}
    \begin{overpic}[width=.162\linewidth]{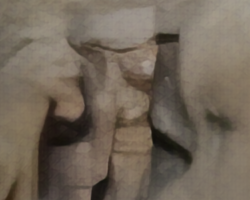}
    \put(2,2){\begin{color}{white}L1=0.005\end{color}}
    \end{overpic}
    \begin{overpic}[width=.162\linewidth]{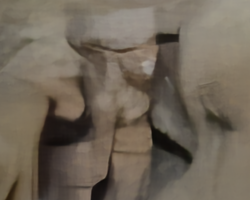}
    \put(2,2){\begin{color}{white}L2=0.001\end{color}}
    \end{overpic}
    \begin{overpic}[width=.162\linewidth]{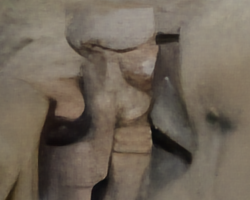}
    \put(2,2){\begin{color}{white}CTXL2=0.01\end{color}}
    \end{overpic}
    \begin{overpic}[width=.162\linewidth]{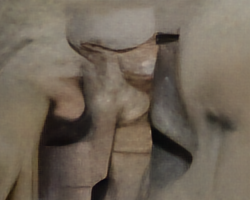}
    \put(2,2){\begin{color}{white}CTXDP=0.01\end{color}}
    \end{overpic}
    \begin{overpic}[width=.162\linewidth]{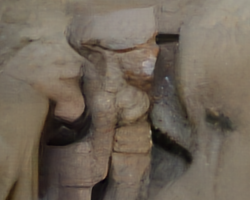}
    \put(2,2){\begin{color}{white}PDL=0.005\end{color}}
    \end{overpic}
    
    \vspace{.20em}
    \begin{overpic}[width=.162\linewidth]{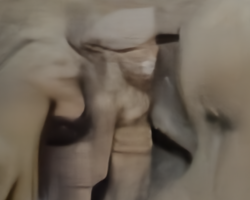}
    \put(2,2){\begin{color}{white}no-perception\end{color}}
    \end{overpic}
    \begin{overpic}[width=.162\linewidth]{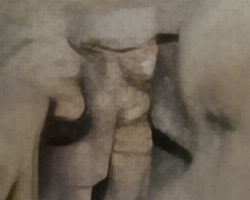}
    \put(2,2){\begin{color}{white}L1=0.01\end{color}}
    \end{overpic}
    \begin{overpic}[width=.162\linewidth]{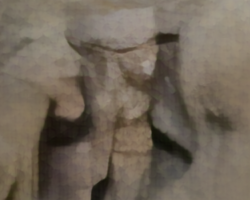}
    \put(2,2){\begin{color}{white}L2=0.01\end{color}}
    \end{overpic}
    \begin{overpic}[width=.162\linewidth]{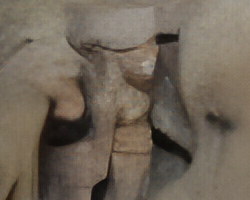}
    \put(2,2){\begin{color}{white}CTXL2=0.1\end{color}}
    \end{overpic}
    \begin{overpic}[width=.162\linewidth]{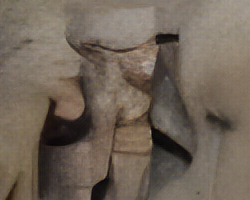}
    \put(2,2){\begin{color}{white}CTXDP=0.1\end{color}}
    \end{overpic}
    \begin{overpic}[width=.162\linewidth]{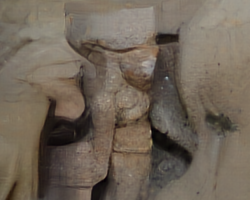}
    \put(2,2){\begin{color}{white}PDL=0.01\end{color}}
    \end{overpic}
    
    \vspace{.20em}
    \begin{overpic}[width=.162\linewidth]{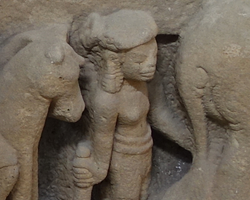}
    \put(2,2){\begin{color}{white}ground-truth\end{color}}
    \end{overpic}
    \begin{overpic}[width=.162\linewidth]{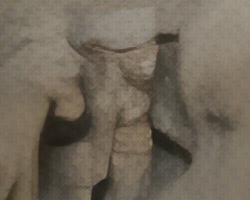}
    \put(2,2){\begin{color}{white}L1=0.1\end{color}}
    \end{overpic}
    \begin{overpic}[width=.162\linewidth]{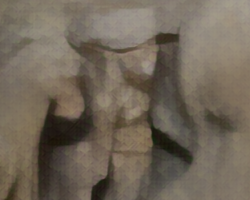}
    \put(2,2){\begin{color}{white}L2=0.05\end{color}}
    \end{overpic}
    \begin{overpic}[width=.162\linewidth]{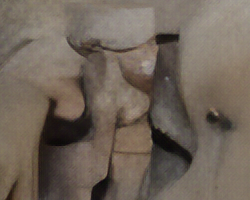}
    \put(2,2){\begin{color}{white}CTXL2=0.5\end{color}}
    \end{overpic}
    \begin{overpic}[width=.162\linewidth]{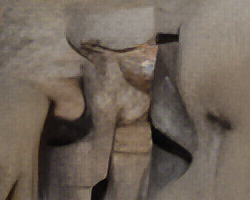}
    \put(2,2){\begin{color}{white}CTXDP=0.5\end{color}}
    \end{overpic}
    \begin{overpic}[width=.162\linewidth]{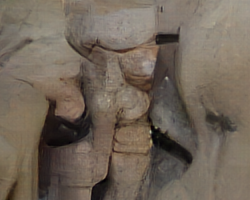}
    \put(2,2){\begin{color}{white}PDL=0.02\end{color}}
    \end{overpic}
    
   \vspace{.2em}
   
    \begin{overpic}[width=.162\linewidth]{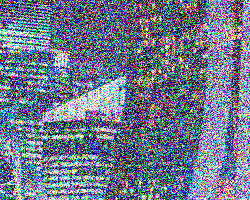}
    \put(2,2){\begin{color}{white}noisy\end{color}}
    \end{overpic}
    \begin{overpic}[width=.162\linewidth]{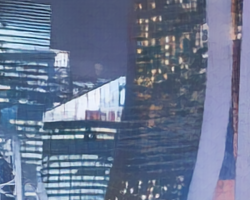}
    \put(2,2){\begin{color}{white}L1=0.05\end{color}}
    \end{overpic}
    \begin{overpic}[width=.162\linewidth]{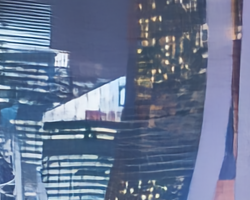}
    \put(2,2){\begin{color}{white}L2=0.001\end{color}}
    \end{overpic}
    \begin{overpic}[width=.162\linewidth]{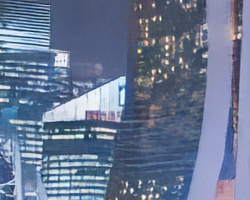}
    \put(2,2){\begin{color}{white}CTXL2=0.01\end{color}}
    \end{overpic}
    \begin{overpic}[width=.162\linewidth]{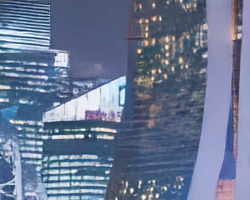}
    \put(2,2){\begin{color}{white}CTXDP=0.01\end{color}}
    \end{overpic}
    \begin{overpic}[width=.162\linewidth]{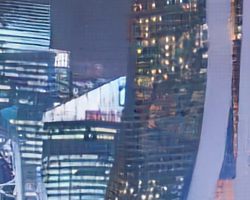}
    \put(2,2){\begin{color}{white}PDL=0.005\end{color}}
    \end{overpic}
    
    \vspace{.20em}
    \begin{overpic}[width=.162\linewidth]{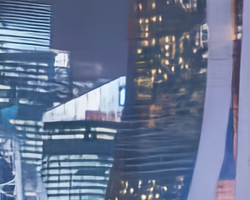}
    \put(2,2){\begin{color}{white}no-perception\end{color}}
    \end{overpic}
    \begin{overpic}[width=.162\linewidth]{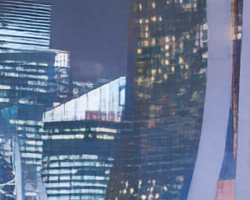}
    \put(2,2){\begin{color}{white}L1=0.01\end{color}}
    \end{overpic}
    \begin{overpic}[width=.162\linewidth]{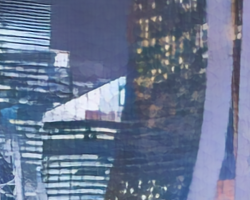}
    \put(2,2){\begin{color}{white}L2=0.01\end{color}}
    \end{overpic}
    \begin{overpic}[width=.162\linewidth]{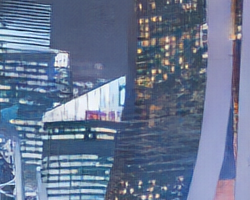}
    \put(2,2){\begin{color}{white}CTXL2=0.1\end{color}}
    \end{overpic}
    \begin{overpic}[width=.162\linewidth]{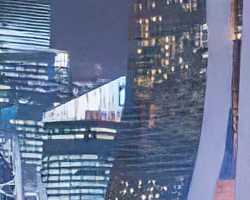}
    \put(2,2){\begin{color}{white}CTXDP=0.1\end{color}}
    \end{overpic}
    \begin{overpic}[width=.162\linewidth]{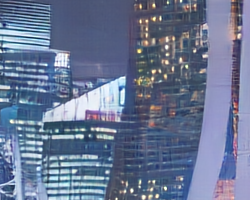}
    \put(2,2){\begin{color}{white}PDL=0.01\end{color}}
    \end{overpic}
    
    \vspace{.20em}
    \begin{overpic}[width=.162\linewidth]{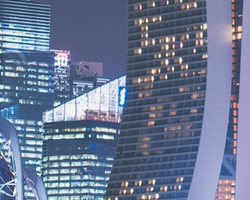}
    \put(2,2){\begin{color}{white}ground-truth\end{color}}
    \end{overpic}
    \begin{overpic}[width=.162\linewidth]{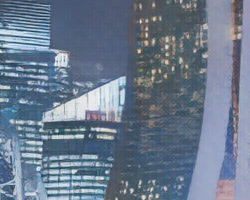}
    \put(2,2){\begin{color}{white}L1=0.1\end{color}}
    \end{overpic}
    \begin{overpic}[width=.162\linewidth]{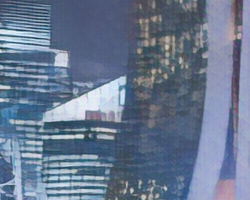}
    \put(2,2){\begin{color}{white}L2=0.05\end{color}}
    \end{overpic}
    \begin{overpic}[width=.162\linewidth]{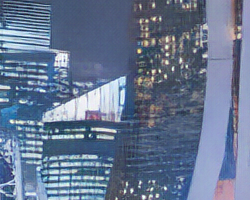}
    \put(2,2){\begin{color}{white}CTXL2=0.5\end{color}}
    \end{overpic}
    \begin{overpic}[width=.162\linewidth]{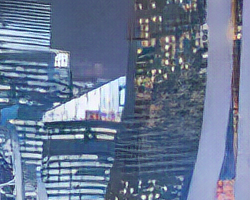}
    \put(2,2){\begin{color}{white}CTXDP=0.5\end{color}}
    \end{overpic}
    \begin{overpic}[width=.162\linewidth]{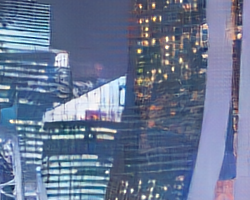}
    \put(2,2){\begin{color}{white}PDL=0.02\end{color}}
    \end{overpic}

    \caption{Denoising at high noise levels $\sigma=100$ example results. For each shown result we indicate the weight value for the perceptual ($L_1$, CTXDP, CTXL2, PDL) loss term.}
    \label{fig:denoise}
\end{figure*}

\vspace{.3em}
\noindent \textbf{Projected Distribution Loss vs other (perceptual) losses.}
Table~\ref{tab:denoising-s100} shows quantitative results of training models using different perceptual losses ($L_1$, $L_2$, Contextual Loss - CTXDP and CTXL2) and no-perceptual loss at all. In Figure~\ref{fig:denoise} we show image crops. 
The first observation is that the results with any perceptual loss are better looking than the one from the model trained without any perceptual loss. This is expected since only penalizing the difference in pixel values leads to blurry images despite having a high PSNR (as shown in Table~\ref{tab:denoising-s100}). 
Directly penalizing the extracted features using the $L_1$ or $L_2$ distance lead to very similar quantitative and qualitative results. The same applies to the two different implementations of the contextual loss (CTXDP and CTXL2). In general, the contextual is the closest comparison to the proposed PDL distribution loss. The PDL results are in general sharper and present better edge and detail structure as shown in the building windows in Figure~\ref{fig:denoise} (middle panel).
In terms of LPIPS the proposed PDL loss and the contextual loss produce similar results for a similar PSNR value. Given that the contextual loss can be seen as an approximation to a distance between distributions, this confirms the superiority of comparing distributions compared to directly comparing the values of the features.
We also carried out a perceptual user study and the average pairwise preferences is presented in Table~\ref{tab:denoising-s100-study}. Overall, PDL outperforms the other models. Our method shows the smallest/largest difference with the no-perceptual/CTXL2 models. Note that performance of the two contextual models are very close to each other.
\begin{table}[h]
\renewcommand{\arraystretch}{0.8}
\setlength{\tabcolsep}{3.5pt}
\footnotesize
    \centering
    \begin{tabular}{rccccccc} \toprule
    
        \multicolumn{2}{l}{Perceptual Loss}	& PSNR	& MS-SSIM  & LPIPS & NIQE & FID  \\\midrule
        \multicolumn{2}{r}{reference}	& -	& - & -	& 3.166 & -  \\
        \multicolumn{2}{r}{input}	& 10.49	& 0.416 &	1.289 &  23.171 & 260.61  \\
        \multicolumn{2}{r}{no-perceptual} &	\textbf{27.14} &	\textbf{0.906} &	0.311 & 3.766 & 84.33 \\\midrule
         & 0.001 &	27.07 &	0.904 &	0.279 & 3.539 & 61.11 \\
         & 0.005 &	26.92 &	0.900 &	0.264 & 4.414 & 59.18  \\
   L1 & 0.010 & 26.81 &	0.896 &	0.268 & 4.456 & 59.89  \\
         & 0.100 &	25.90 &	0.877 &	0.261 & 4.570 & 60.98 \\
         & 0.150 &	25.50 &	0.871 &	0.270 & 4.799 & 62.57 \\\midrule
         & 0.001 &  27.02 & 0.905 & 0.283 & 3.639 & 69.38  \\
  L2  & 0.010 &	26.72 &	0.895 &	0.296 & 3.964 & 64.74 \\
         & 0.050 &	26.30 &	0.884 &	0.308 & 5.381 & 72.55 \\\midrule
         & 0.010 &	26.91 &	0.902 &	0.252 & 3.402 & 59.28 \\
  CTXDP  & 0.100 &  26.57 &	0.896 &	0.241 & 3.490 & 58.82  \\
         & 0.500 & 	26.34 &	0.894 &	0.249 & 3.480 & 60.98  \\\midrule
         & 0.010 &	26.85 &	0.901 &	0.246 & 3.510 & 61.60 \\
   CTXL2 & 0.100 &	26.44 &	0.896 &	0.239 & 3.740 & 57.92 \\
         & 0.500 &	25.98 &	0.891 &	0.244 & 3.404 & 56.32 \\ \midrule
         & 0.001 &	27.10 &	0.906 &	0.250 & \textbf{3.226} & 57.47 \\
         & 0.005 &	26.74 &	0.899 &	0.243 & 3.291 & 55.26 \\
    PDL  & 0.010 &	26.62 & 0.898 & \textbf{0.233} & 3.398 & \textbf{54.51}  \\
         & 0.015 &	26.48 &	0.895 &	0.238 & 3.275 & 54.95  \\
         & 0.020 &	26.36 &	0.893 &	0.239 & 3.483 & 58.12 \\\midrule
    \end{tabular}
    \vspace{.2em}
 \caption{Average performance metrics for Denoising at high noise levels $\sigma=100$. All models were trained using the same model configuration and optimization parameters. Results with different weight for the respective perceptual loss term are presented. All perceptual looses are computed on \texttt{VGG16\-conv4} features. The best results are highlighted in bold.}    \label{tab:denoising-s100}
\end{table}

\begin{table}[t]
\renewcommand{\arraystretch}{0.8}
\footnotesize
    \centering
    \begin{tabular}{cccccc} \toprule
             & no-perceptual & L1	& CTXDP	& CTXL2 & PDL \\\midrule
     no-perceptual & - & 0.43& 0.31	 & 0.29	 & 0.23 \\
      L1 & 0.57 & - & 0.40 & 0.38 & 0.26 \\
    CTXDP  & 0.69 &	0.60 & - & 0.47	 & 0.39 \\
     CTXL2 & 0.71 &	0.62 &	0.53 & - & 0.43 \\
     PDL & \textbf{0.77} & \textbf{0.74} & \textbf{0.61} &	\textbf{0.57} & - \\\midrule
    \end{tabular}
    \vspace{.2em}
    \caption{Average pairwise human preference for denoising at high noise levels $\sigma=100$. Each value represents the fraction of times the Amazon Mechanical Turk raters chose the row over the column. The average number of raters is 25. All models were trained using the same model configuration and optimization parameters. Results with the best loss weight are shown here. The best results are highlighted in bold.}
    \label{tab:denoising-s100-study}
\end{table}

\subsection{Single-Image Super-resolution}
\begin{figure*}
    \centering
    \scriptsize
    
    \begin{minipage}[c]{.23\linewidth}
    \begin{tikzpicture}
      \node[anchor=north west,inner sep=0] (image) at (0,0) {\includegraphics[clip, trim=600 800 550 430, width=\linewidth]{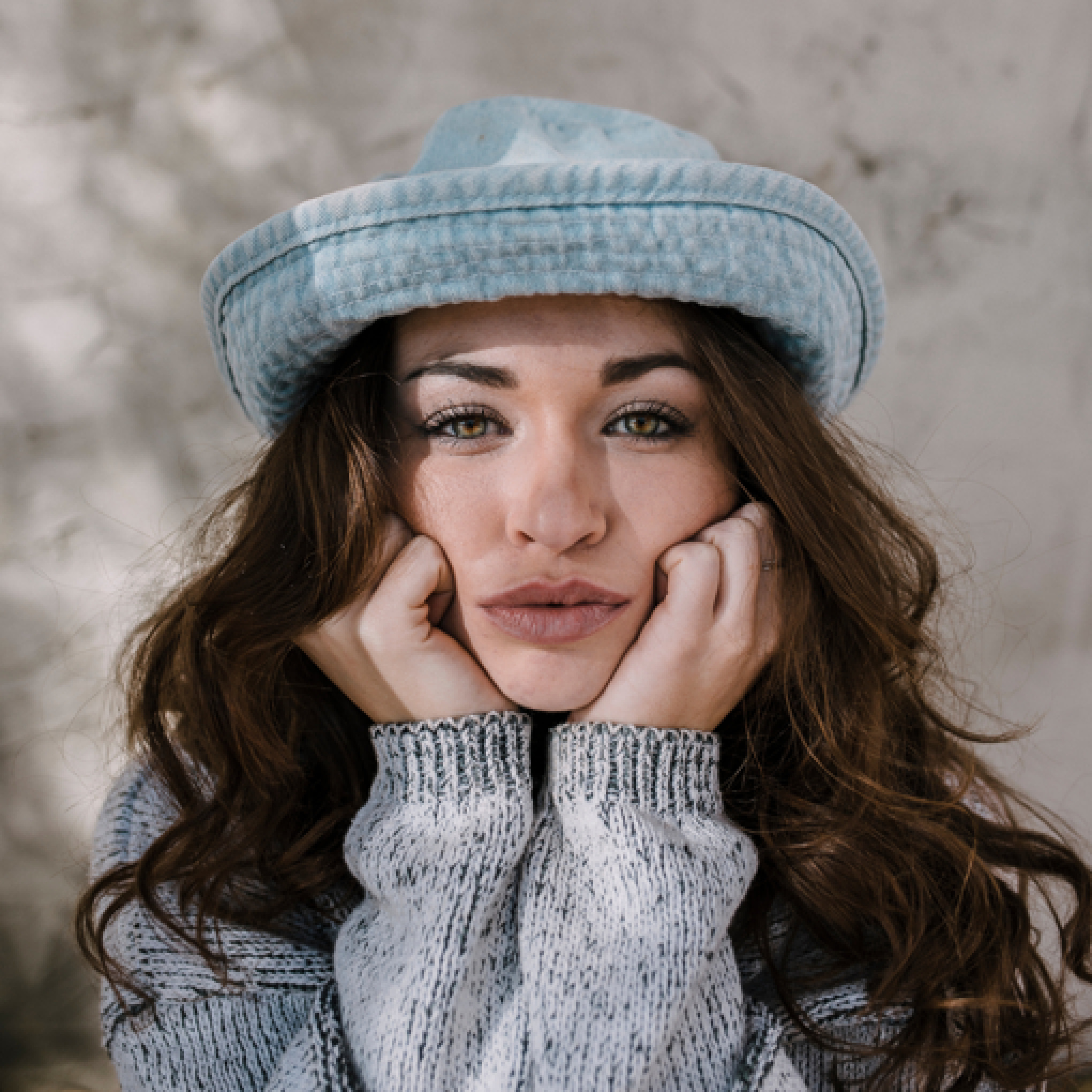} };
      \begin{scope}[x={(image.north east)},y={(image.south west)}]
        \draw[red1, very thick] (0.5506, 0.3580) rectangle (0.8652, 0.5556);
        \node[] at (0.12,0.955) {\begin{color}{white}low-res\end{color}};
      \end{scope}
     \end{tikzpicture} \vspace{-.9em}
     
      \begin{tikzpicture}
      \node[anchor=north west,inner sep=0] (image) at (0,0) {\includegraphics[clip, trim=600 800 550 430, width=\linewidth]{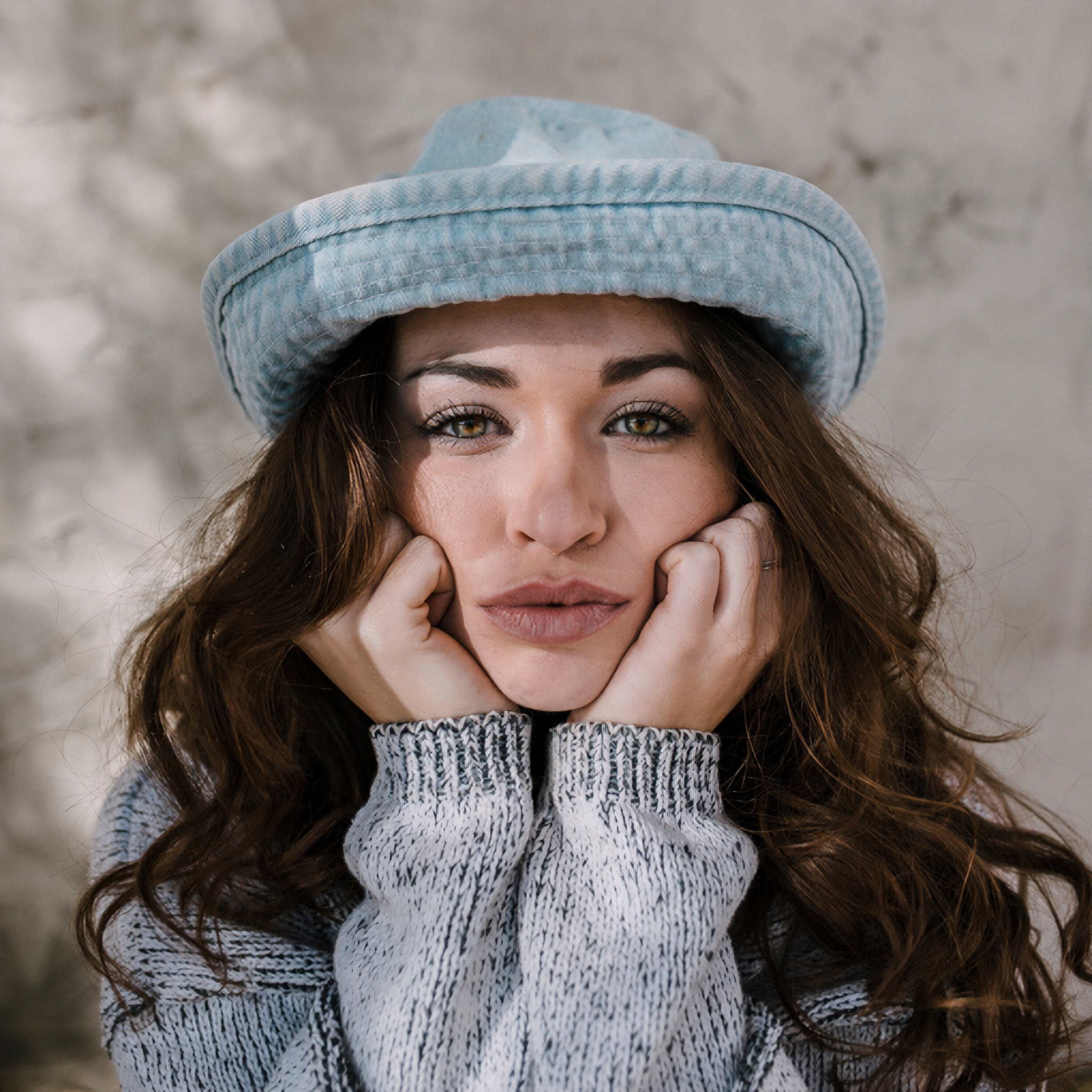} };
      \begin{scope}[x={(image.north east)},y={(image.south west)}]
        \draw[red1, very thick] (0.5506, 0.3580) rectangle (0.8652, 0.5556);
        \node[] at (0.15,0.955) {\begin{color}{white}PDL=0.01\end{color}};
      \end{scope}
     \end{tikzpicture}  
    
    \end{minipage}
    \vspace{.2em}
    \begin{minipage}[c]{.74\linewidth} 
    \begin{overpic}[width=.325\linewidth]{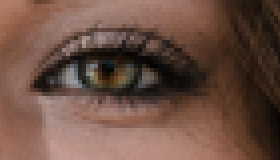}
    \put(2,2){\begin{color}{white}low-res\end{color}}
    \end{overpic}
    \begin{overpic}[width=.325\linewidth]{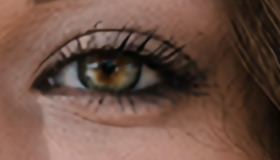}
    \put(2,2){\begin{color}{white}no-perceptual\end{color}}
    \end{overpic}
    \begin{overpic}[width=.325\linewidth]{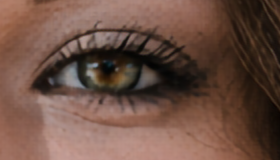}
    \put(2,2){\begin{color}{white}L1=0.01\end{color}}
    \end{overpic} \vspace{.2em}
    
    \begin{overpic}[width=.325\linewidth]{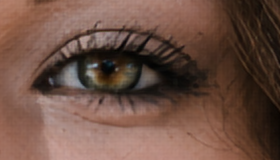}
    \put(2,2){\begin{color}{white}CTXDP=0.01\end{color}}
    \end{overpic}
    \begin{overpic}[width=.325\linewidth]{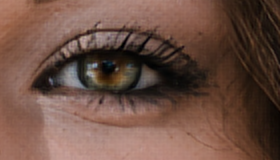}
    \put(2,2){\begin{color}{white}CTXL2=0.01\end{color}}
    \end{overpic}
    \begin{overpic}[width=.325\linewidth]{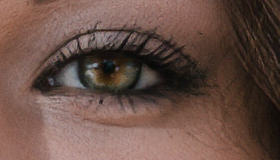}
    \put(2,2){\begin{color}{white}SRGAN\end{color}}
    \end{overpic} \vspace{.2em}
    
    \begin{overpic}[width=.325\linewidth]{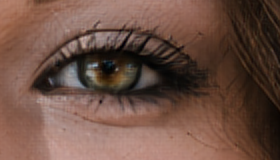}
    \put(2,2){\begin{color}{white}PDL=0.003\end{color}}
    \end{overpic}
    \begin{overpic}[width=.325\linewidth]{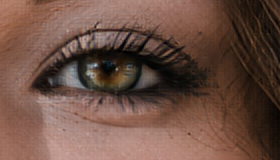}
    \put(2,2){\begin{color}{white}PDL=0.01\end{color}}
    \end{overpic}
    \begin{overpic}[width=.325\linewidth]{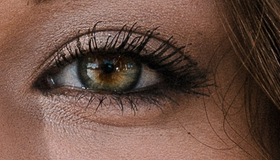}
    \put(2,2){\begin{color}{white}ground-truth\end{color}}
    \end{overpic}
    \end{minipage}

    \begin{minipage}[c]{.23\linewidth}
    \begin{tikzpicture}
      \node[anchor=north west,inner sep=0] (image) at (0,0) {\includegraphics[clip, trim=550 600 1050 360, width=\linewidth]{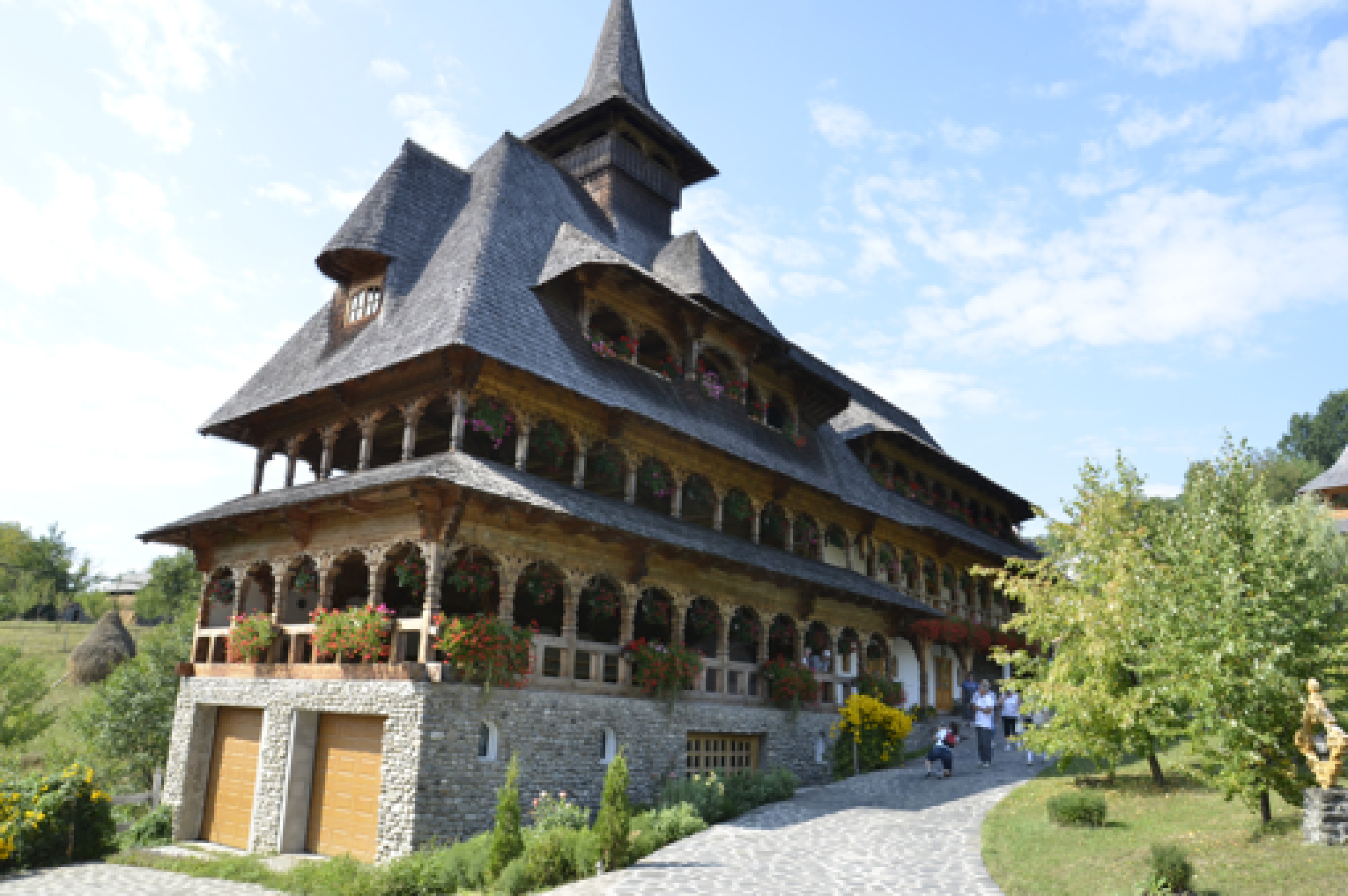} };
      \begin{scope}[x={(image.north east)},y={(image.south west)}]
        \draw[yellow1, very thick] (0.3729, 0.2512) rectangle (0.7525, 0.5516);
        \node[] at (0.12,0.955) {\begin{color}{white}low-res\end{color}};
      \end{scope}
     \end{tikzpicture} 
     
      \begin{tikzpicture}
      \node[anchor=north west,inner sep=0] (image) at (0,0) {\includegraphics[clip, trim=550 600 1050 360, width=\linewidth]{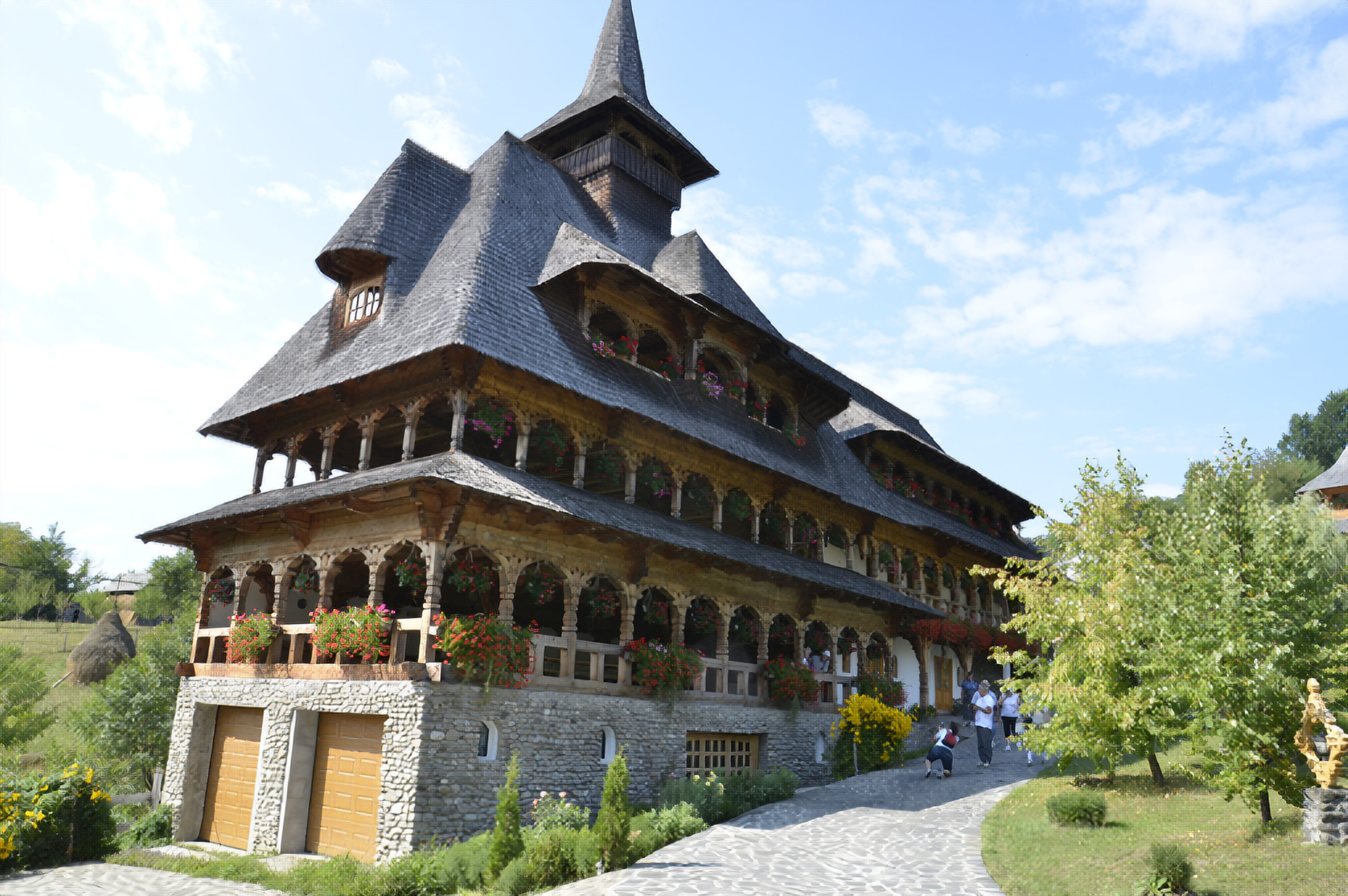} };
      \begin{scope}[x={(image.north east)},y={(image.south west)}]
        \draw[yellow1, very thick] (0.3729, 0.2512) rectangle (0.7525, 0.5516);
        \node[] at (0.15,0.955) {\begin{color}{white}PDL=0.01\end{color}};
      \end{scope}
     \end{tikzpicture}  
    
    \end{minipage}
    \vspace{.2em}
     \begin{minipage}[c]{.74\linewidth}
    \begin{overpic}[width=.325\linewidth]{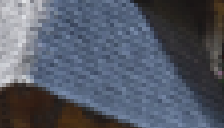}
    \put(2,2){\begin{color}{white}low-res\end{color}}
    \end{overpic}
    \begin{overpic}[width=.325\linewidth]{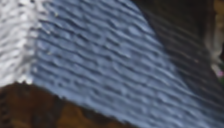}
    \put(2,2){\begin{color}{white}no-perceptual\end{color}}
    \end{overpic}
    \begin{overpic}[width=.325\linewidth]{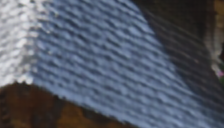}
    \put(2,2){\begin{color}{white}L1=0.01\end{color}}
    \end{overpic} \vspace{.2em}
    
    \begin{overpic}[width=.325\linewidth]{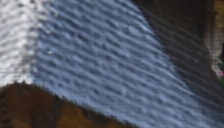}
    \put(2,2){\begin{color}{white}CTXDP=0.01\end{color}}
    \end{overpic}
    \begin{overpic}[width=.325\linewidth]{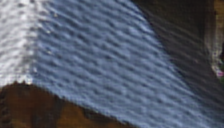}
    \put(2,2){\begin{color}{white}CTXL2=0.01\end{color}}
    \end{overpic}
    \begin{overpic}[width=.325\linewidth]{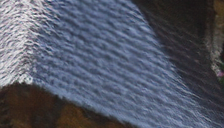}
    \put(2,2){\begin{color}{white}SRGAN\end{color}}
    \end{overpic} \vspace{.2em}
    
    \begin{overpic}[width=.325\linewidth]{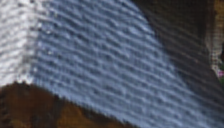}
    \put(2,2){\begin{color}{white}PDL=0.003\end{color}}
    \end{overpic}
    \begin{overpic}[width=.325\linewidth]{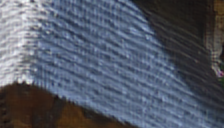}
    \put(2,2){\begin{color}{white}PDL=0.01\end{color}}
    \end{overpic}
    \begin{overpic}[width=.325\linewidth]{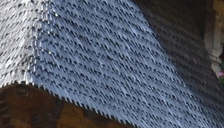}
    \put(2,2){\begin{color}{white}ground-truth\end{color}}
    \end{overpic}
   \end{minipage}

    \caption{Single image $\times 4$ super-resolution example results. For each shown result we indicate the weight value for the perceptual ($L_1$, CTXDP, CTXL2, PDL) loss term.}
    \label{fig:sisr}
\end{figure*}

We evaluated our proposed loss on single-image $4\times$ super-resolution. In addition to comparing to the different losses we already mentioned we include a comparison with SRGAN~\cite{ledig2017photo}.
Table~\ref{tab:sisr-4x} summarizes the quantitative results of the different evaluated losses on $4\times$ super-resolution using the div2k training and validation dataset (bicubic downscaling). A selection of results is presented in Figure~\ref{fig:sisr}. Our proposed PDL loss leads to super-resolved images with more defined structure. SRGAN produces more fine grain details but also tends to hallucinate more content (for instance the structure on the roof). This is also reflected in the low PSNR score implying a significant pixel distortion.
Results from our subjective study are shown in Table~\ref{tab:sisr-4x-study}. We observe that on average PDL and SRGAN perform better than the other compared methods.

\begin{table}[h]
\renewcommand{\arraystretch}{0.8}
\setlength{\tabcolsep}{3.5pt}
\footnotesize
    \centering
    \begin{tabular}{lccccc} \toprule
        Perceptual Loss	& PSNR	& MS-SSIM & LPIPS & NIQE & FID  \\\midrule
         reference     &	-     &	-     &	-     & 3.166 & -  \\
         no-perceptual &	\textbf{28.76} &	\textbf{0.966} &	0.274 & 4.762 & 25.30 \\
                    L1 &	28.23 &	0.961 &	0.161 & 4.914 & \textbf{16.37} \\
                 CTXDP &    28.53 & 0.964 &	0.188 & 4.340 & 18.40  \\
                 CTXL2 &	28.40 &	0.963 &	0.173 & 4.169 & 17.92  \\
                 SRGAN &	26.21 &	0.941 &	0.150 & \textbf{2.975} & 18.05  \\
PDL ($\lambda=0.003$)  &    28.21 &	0.961 &	0.169 & 3.892 & 18.15  \\
PDL ($\lambda=0.01$)   &	27.81 &	0.957 &	\textbf{0.145} & 3.989 & 16.66  \\\midrule
    \end{tabular}
    \vspace{.2em}
    \caption{Single image $4\times$ super-resolution. All models were trained using the same model configuration and optimization parameters. The best results are highlighted in bold.}   
    \label{tab:sisr-4x}
\end{table}

\begin{table}[t]
\renewcommand{\arraystretch}{0.8}
\footnotesize
    \centering
    \begin{tabular}{ccccccc} \toprule
             & no-perc & L1	& CTXDP	& CTXL2 & SRGAN & PDL \\\midrule
     no-perceptual & - & 0.43 & 0.31 & 0.30 & 0.28  & 0.24\\
     L1 & 0.57 & - & 0.41 & 0.42 & 0.37 & 0.34\\
     CTXDP  & 0.69 & 0.59 & - & 0.47 & 0.41 & 0.38\\
     CTXL2 & 0.70 & 0.58 & 0.53 & - & 0.42  & 0.39\\
     SRGAN & 0.72 & 0.63 & 0.59 & 0.58 & -  & \textbf{0.50}\\
     PDL & \textbf{0.76} & \textbf{0.66} & \textbf{0.62} & \textbf{0.61} & \textbf{0.50} & -\\\midrule
    \end{tabular}
    \vspace{.2em}
    \caption{Average pairwise human preference for single image 4x super-resolution. Each value represents the fraction of times the Amazon Mechanical Turk raters chose the row over the column. The average number of raters is 25. Results with the best loss weight are shown here. The best results are highlighted in bold.}
    \label{tab:sisr-4x-study}
\end{table}

\subsection{Deblurring}
\begin{figure*}
    \centering
    \scriptsize

    \begin{minipage}[c]{.33\linewidth}
    \begin{tikzpicture}
      \node[anchor=north west,inner sep=0] (image) at (0,0) {\includegraphics[clip, trim=2 150 0 50, width=\linewidth]{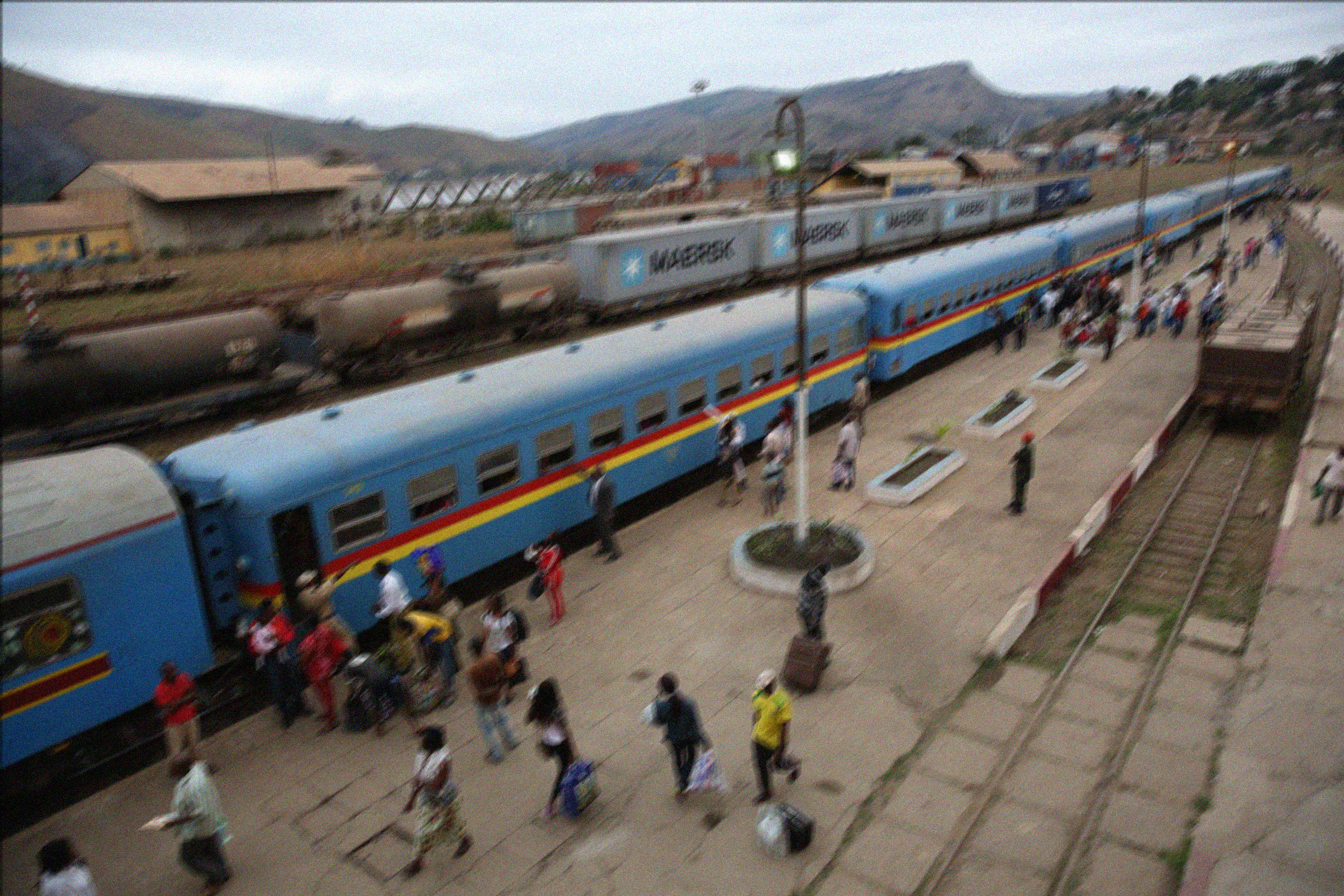} };
      \begin{scope}[x={(image.north east)},y={(image.south west)}]
        \draw[yellow1, very thick] (0.3760, 0.1582) rectangle (0.4752, 0.2981);
        \node[] at (0.12,0.955) {\begin{color}{white}blurry-noisy\end{color}};
      \end{scope}
     \end{tikzpicture} \vspace{-.9em}
     
      \begin{tikzpicture}
      \node[anchor=north west,inner sep=0] (image) at (0,0) {\includegraphics[clip, trim=2 150 0 50, width=\linewidth]{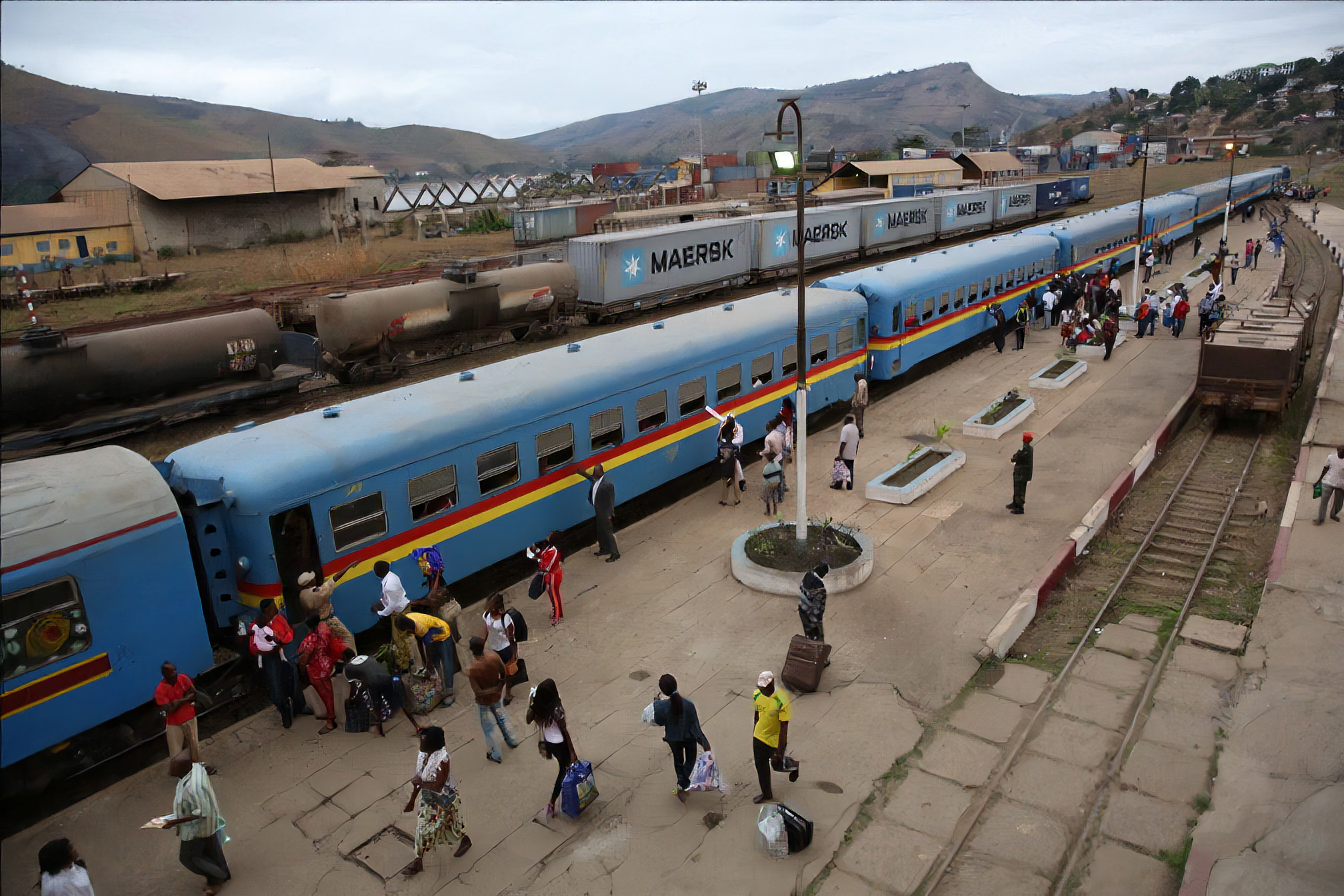} };
      \begin{scope}[x={(image.north east)},y={(image.south west)}]
        \draw[yellow1, very thick] (0.3760, 0.1582) rectangle (0.4752, 0.2981);
        \node[] at (0.05,0.955) {\begin{color}{white}\textsc{PDL}\end{color}};
      \end{scope}
     \end{tikzpicture}  
     \end{minipage} 
     \begin{minipage}[c]{.33\linewidth}
    \begin{tikzpicture}
      \node[anchor=north west,inner sep=0] (image) at (0,0) {\includegraphics[clip, trim=10 180 0 50, width=\linewidth]{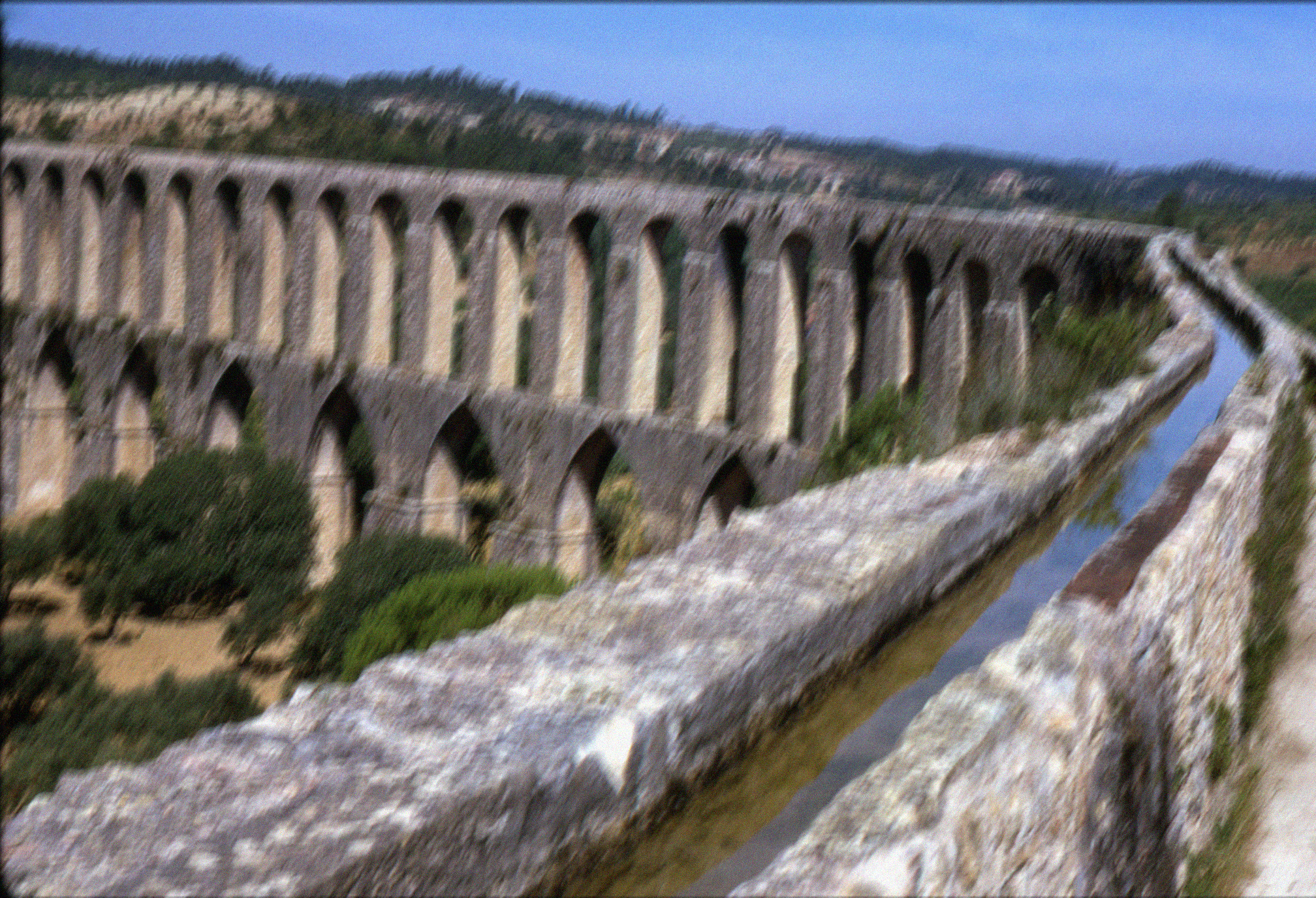} };
      \begin{scope}[x={(image.north east)},y={(image.south west)}]
        \draw[red1, very thick] (0.6250, 0.4165) rectangle (0.7242, 0.5586);
        \node[] at (0.12,0.955) {\begin{color}{white}blurry-noisy\end{color}};
      \end{scope}
     \end{tikzpicture} \vspace{-.9em}
     
      \begin{tikzpicture}
      \node[anchor=north west,inner sep=0] (image) at (0,0) {\includegraphics[clip, trim=10 180 0 50, width=\linewidth]{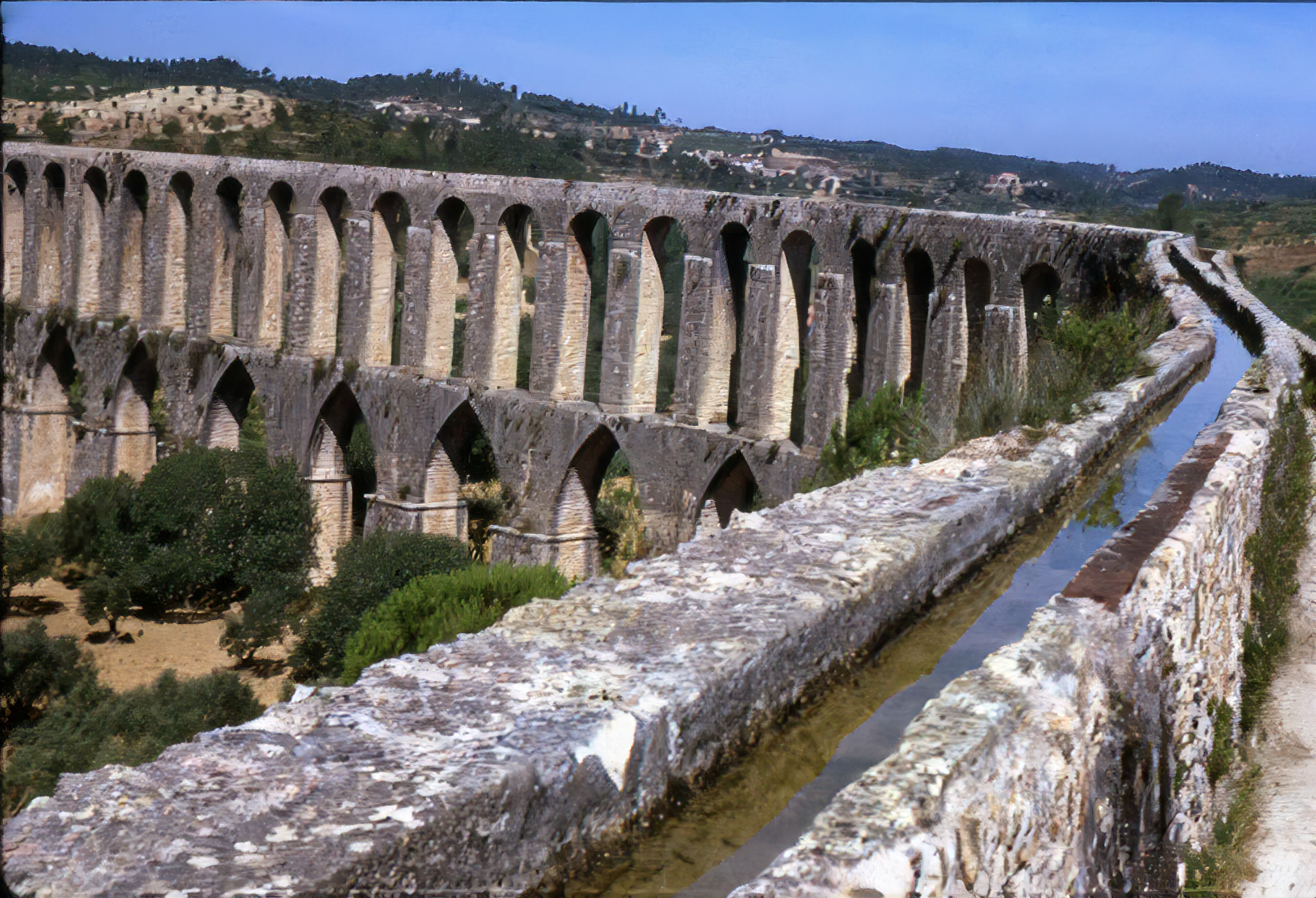} };
      \begin{scope}[x={(image.north east)},y={(image.south west)}]
         \draw[red1, very thick] (0.6250, 0.4165) rectangle (0.7242, 0.5586);
        \node[] at (0.05,0.955) {\begin{color}{white}\textsc{PDL}\end{color}};
      \end{scope}
     \end{tikzpicture}  
     \end{minipage} 
    \begin{minipage}[c]{.33\linewidth}
    \begin{tikzpicture}
      \node[anchor=north west,inner sep=0] (image) at (0,0) {\includegraphics[clip, trim=10 100 0 100, width=\linewidth]{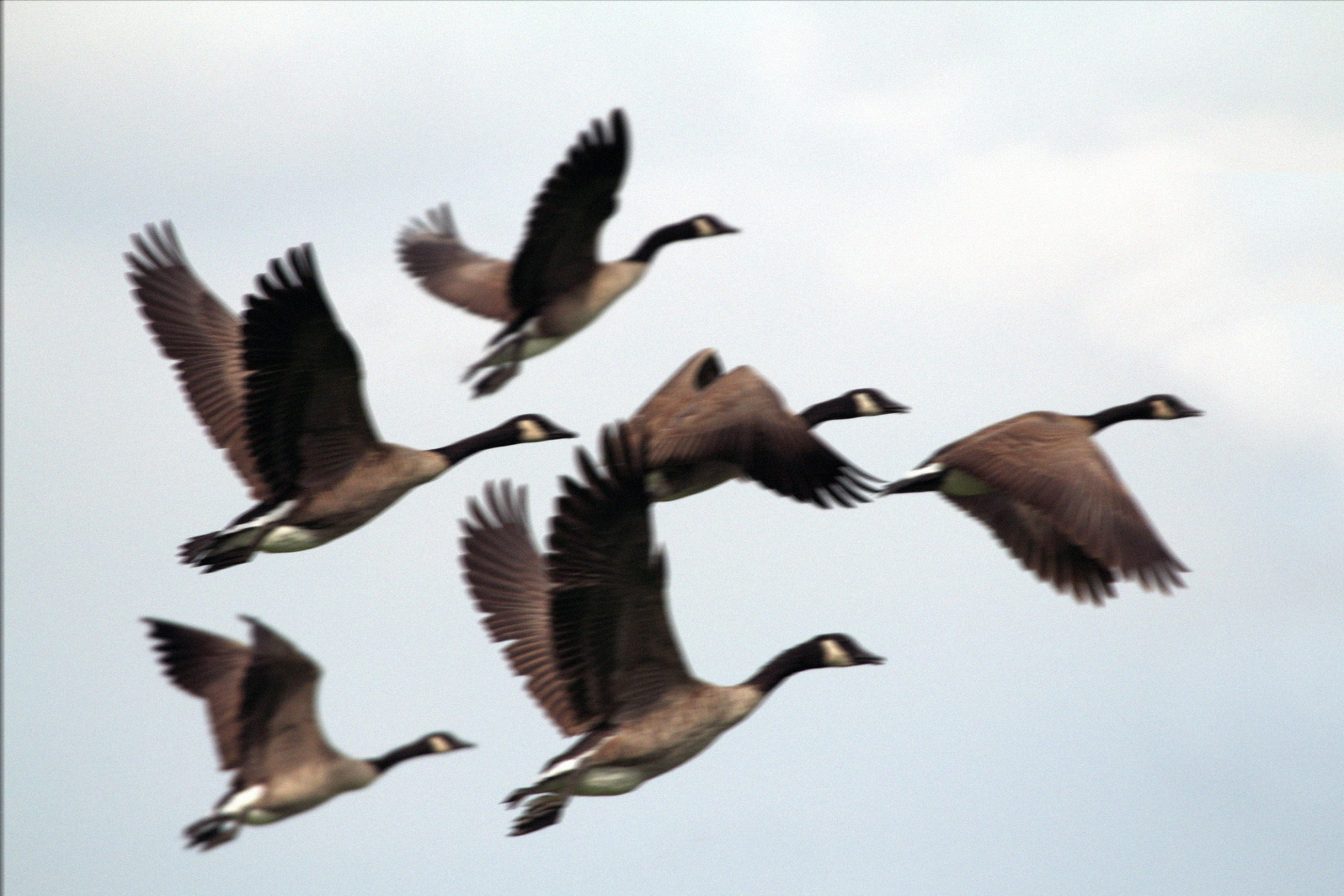} };
      \begin{scope}[x={(image.north east)},y={(image.south west)}]
         \draw[blue1, very thick] (0.2418, 0.4502) rectangle (0.3415, 0.5900);
         \node[] at (0.12,0.955) {\begin{color}{white}blurry-noisy\end{color}};
      \end{scope}
     \end{tikzpicture} \vspace{-.9em}
     
      \begin{tikzpicture}
      \node[anchor=north west,inner sep=0] (image) at (0,0) {\includegraphics[clip, trim=10 100 0 100, width=\linewidth]{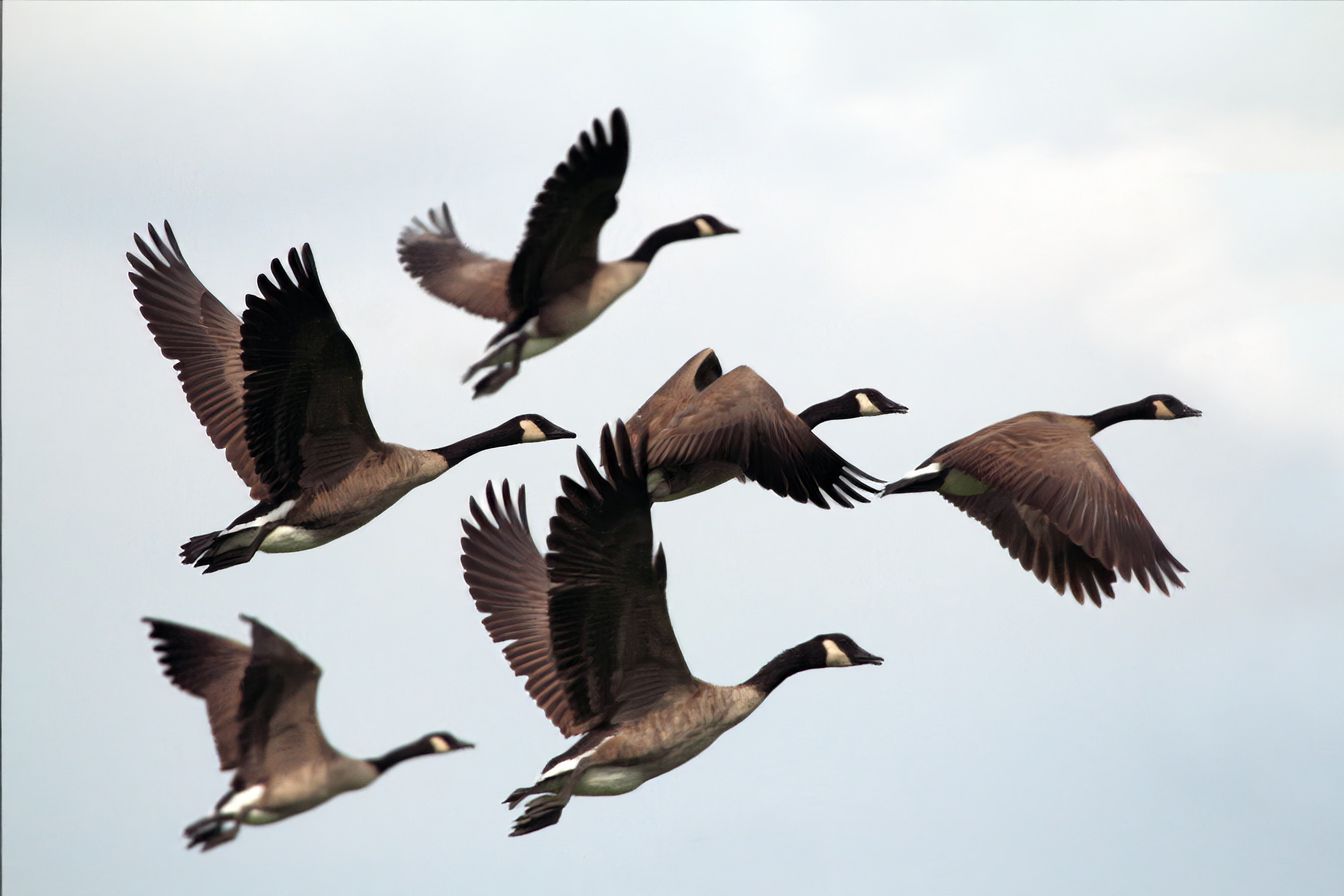} };
      \begin{scope}[x={(image.north east)},y={(image.south west)}]
         \draw[blue1, very thick] (0.2418, 0.4502) rectangle (0.3415, 0.5900);
         \node[] at (0.05,0.955) {\begin{color}{white}\textsc{PDL}\end{color}};
      \end{scope}
     \end{tikzpicture}  
     \end{minipage} 
     
     \vspace{.2em}
     
    \begin{overpic}[width=.139\linewidth]{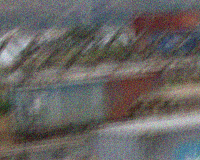}
    \put(2,2){\begin{color}{white}blurry-noisy\end{color}}
    \end{overpic}
    \begin{overpic}[width=.139\linewidth]{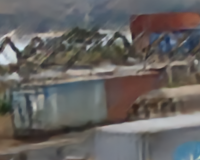}
    \put(2,2){\begin{color}{white}no-perceptual\end{color}}
    \end{overpic}
    \begin{overpic}[width=.139\linewidth]{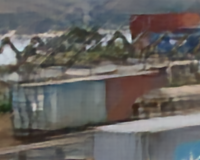}
    \put(2,2){\begin{color}{white}L1=0.01\end{color}}
    \end{overpic}
    \begin{overpic}[width=.139\linewidth]{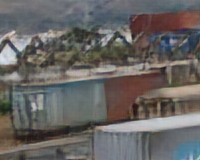}
    \put(2,2){\begin{color}{white}CTXDP=0.01\end{color}}
    \end{overpic}
    \begin{overpic}[width=.139\linewidth]{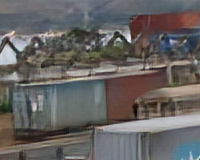}
    \put(2,2){\begin{color}{white}CTXL2=0.01\end{color}}
    \end{overpic}
    \begin{overpic}[width=.139\linewidth]{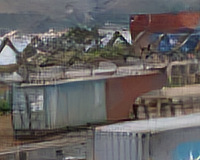}
    \put(2,2){\begin{color}{white}PDL=0.01\end{color}}
    \end{overpic}
    \begin{overpic}[width=.139\linewidth]{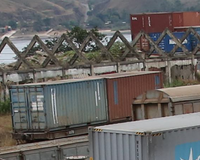}
    \put(2,2){\begin{color}{white}ground-truth\end{color}}
    \end{overpic}
    
    \vspace{.2em}

    \begin{overpic}[width=.139\linewidth]{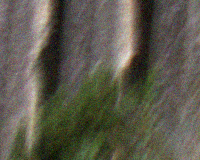}
    \put(2,2){\begin{color}{white}blurry-noisy\end{color}}
    \end{overpic}
    \begin{overpic}[width=.139\linewidth]{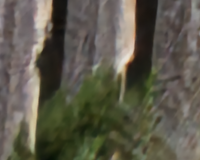}
    \put(2,2){\begin{color}{white}no-perceptual\end{color}}
    \end{overpic}
    \begin{overpic}[width=.139\linewidth]{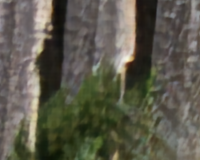}
    \put(2,2){\begin{color}{white}L1=0.01\end{color}}
    \end{overpic}
    \begin{overpic}[width=.139\linewidth]{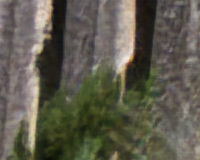}
    \put(2,2){\begin{color}{white}CTXDP=0.01\end{color}}
    \end{overpic}
    \begin{overpic}[width=.139\linewidth]{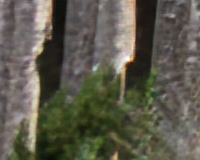}
    \put(2,2){\begin{color}{white}CTXL2=0.01\end{color}}
    \end{overpic}
    \begin{overpic}[width=.139\linewidth]{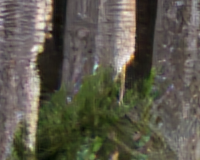}
    \put(2,2){\begin{color}{white}PDL=0.01\end{color}}
    \end{overpic}
    \begin{overpic}[width=.139\linewidth]{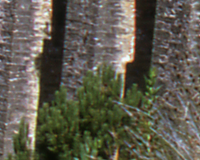}
    \put(2,2){\begin{color}{white}ground-truth\end{color}}
    \end{overpic}

    \vspace{.2em}

    \begin{overpic}[width=.139\linewidth]{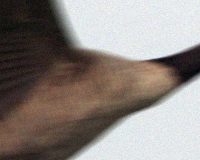}
    \put(2,2){\begin{color}{white}blurry-noisy\end{color}}
    \end{overpic}
    \begin{overpic}[width=.139\linewidth]{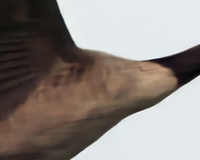}
    \put(2,2){\begin{color}{white}no-perceptual\end{color}}
    \end{overpic}
    \begin{overpic}[width=.139\linewidth]{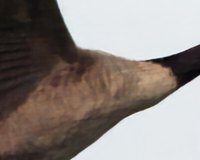}
    \put(2,2){\begin{color}{white}L1=0.01\end{color}}
    \end{overpic}
    \begin{overpic}[width=.139\linewidth]{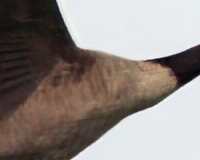}
    \put(2,2){\begin{color}{white}CTXDP=0.01\end{color}}
    \end{overpic}
    \begin{overpic}[width=.139\linewidth]{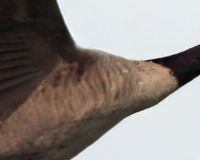}
    \put(2,2){\begin{color}{white}CTXL2=0.01\end{color}}
    \end{overpic}
    \begin{overpic}[width=.139\linewidth]{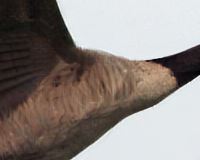}
    \put(2,2){\begin{color}{white}PDL=0.01\end{color}}
    \end{overpic}
    \begin{overpic}[width=.139\linewidth]{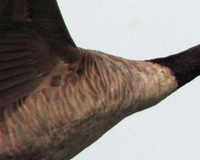}
    \put(2,2){\begin{color}{white}ground-truth\end{color}}
    \end{overpic}

    \vspace{.2em}

    \caption{Deblurring example results. For each shown result we indicate the weight value for the perceptual ($L_1$, CTXDP, CTXL2, PDL) loss term.}
    \label{fig:deblurring}
\end{figure*}

In order to evaluate the performance of the proposed loss on a deblurring task we simulated motion blur due to camera shake. Following~\cite{delbracio2015burst}, we simulated random blur kernels mimicking camera shake blur of varying intensity ($31\times31$ maximal support), and also added random additive noise. The added noise is also of random noise level ($\sigma \in [0,15]$). We trained the DsDeblur~\cite{gao2019dynamic} model using our proposed PDL loss and all the previously discussed losses. Table~\ref{tab:deblurring} summarizes the average quantiative performance of all tested models. As can be seen, our proposed PDL loss produces high-quality results both in term of PSNR and LPIPS.  In Figure~\ref{fig:deblurring} we show several example results.

\begin{table}[t]
\renewcommand{\arraystretch}{0.8}
\setlength{\tabcolsep}{3.5pt}
\footnotesize
    \centering
    \begin{tabular}{lccccc} \toprule
        Perceptual Loss	& PSNR	& MS-SSIM & LPIPS & NIQE & FID \\\midrule
        reference      &	-     &	-     &	-     & 3.166 & -  \\
                 input &	27.07 &	0.899 &	0.328 & 7.482 & 45.96 \\
         no-perceptual &	32.57 &	0.969 &	0.161 & 4.009 & 22.99 \\
                    L1 &	\textbf{33.07} &	\textbf{0.973} & 0.126 & 3.768 & 14.29 \\
                 CTXDP &    32.11 & 0.965 &	0.117 & 3.666 & 15.60 \\
                 CTXL2 &	32.79 &	0.971 &	0.095 & 3.635 & 12.71 \\
PDL ($\lambda=0.001$)  &    33.01 &	0.974 &	0.112 & 3.546 & 13.76 \\
PDL ($\lambda=0.005$)  &	32.53 &	0.969 &	\textbf{0.092} & \textbf{3.476} & \textbf{11.76} \\
PDL ($\lambda=0.010$)   &	31.97 &	0.964 &	0.093 & 3.572 & 13.36 \\\midrule
    \end{tabular}
    \vspace{.2em}
    \caption{Camera shake deblurring. Best results are highlighted in bold.}   
    \label{tab:deblurring}
\end{table}

\subsection{Other applications}

\begin{figure}
    \centering
    
    \scriptsize
    
    \begin{minipage}[c]{\linewidth}
    \begin{overpic}[clip, trim=0 50 0 0, width=.2425\linewidth]{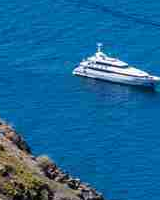}
    \put(3,4){\begin{color}{white}jpeg-q20\end{color}}
    \end{overpic}
    \begin{overpic}[clip, trim=0 50 0 0, width=.2425\linewidth]{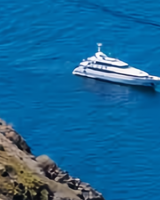}
    \put(3,4){\begin{color}{white}no-perceptual\end{color}}
    \end{overpic}
    \begin{overpic}[clip, trim=0 50 0 0, width=.2425\linewidth]{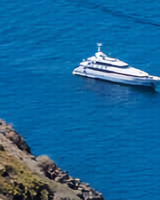}
    \put(3,4){\begin{color}{white}L1=0.01\end{color}}
    \end{overpic}
    \begin{overpic}[clip, trim=0 50 0 0, width=.2425\linewidth]{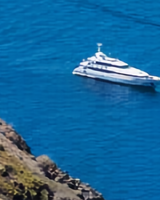}
    \put(3,4){\begin{color}{white}L2=0.01\end{color}}
    \end{overpic}
    \end{minipage} \vspace{.25em}
    
    \begin{minipage}[c]{\linewidth}
    \begin{overpic}[clip, trim=0 50 0 0, width=.2425\linewidth]{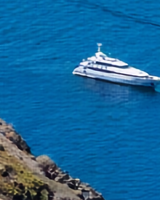}
    \put(3,4){\begin{color}{white}CTXDP=0.01\end{color}}
    \end{overpic}
    \begin{overpic}[clip, trim=0 50 0 0, width=.2425\linewidth]{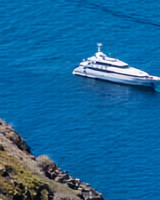}
    \put(3,4){\begin{color}{white}CTXL2=0.01\end{color}}
    \end{overpic}
    \begin{overpic}[clip, trim=0 50 0 0, width=.2425\linewidth]{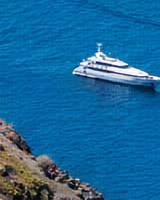}
    \put(3,4){\begin{color}{white}PDL=0.01\end{color}}
    \end{overpic}
    \begin{overpic}[clip, trim=0 50 0 0, width=.2425\linewidth]{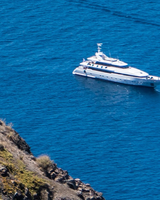}
    \put(3,4){\begin{color}{white}ground-truth\end{color}}
    \end{overpic}
    
    \end{minipage}\vspace{.25em}
    
    \begin{minipage}[c]{\linewidth}
    \begin{overpic}[clip, trim=0 40 0 10, width=.2425\linewidth]{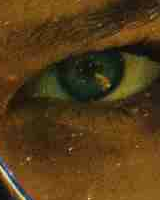}
    \put(4,4){\begin{color}{white}jpeg-q20\end{color}}
    \end{overpic}
    \begin{overpic}[clip, trim=0 40 0 10, width=.2425\linewidth]{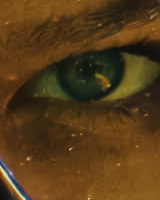}
    \put(4,4){\begin{color}{white}no-perceptual\end{color}}
    \end{overpic}
    \begin{overpic}[clip, trim=0 40 0 10, width=.2425\linewidth]{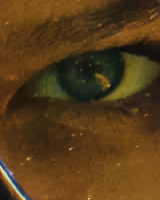}
    \put(4,4){\begin{color}{white}L1=0.01\end{color}}
    \end{overpic}
    \begin{overpic}[clip, trim=0 40 0 10, width=.2425\linewidth]{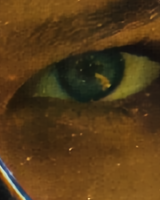}
    \put(4,4){\begin{color}{white}L2=0.01\end{color}}
    \end{overpic} 
    \end{minipage} \vspace{.2em}
    
    \begin{minipage}[c]{\linewidth}
    \begin{overpic}[clip, trim=0 40 0 10, width=.2425\linewidth]{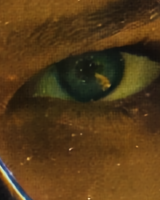}
    \put(4,4){\begin{color}{white}CTXDP=0.01\end{color}}
    \end{overpic}
    \begin{overpic}[clip, trim=0 40 0 10, width=.2425\linewidth]{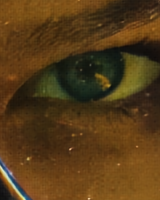}
    \put(4,4){\begin{color}{white}CTXL2=0.01\end{color}}
    \end{overpic}
    \begin{overpic}[clip, trim=0 40 0 10, width=.2425\linewidth]{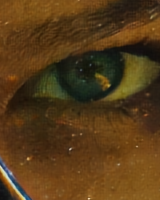}
    \put(4,4){\begin{color}{white}PDL=0.01\end{color}}
    \end{overpic}
    \begin{overpic}[clip, trim=0 40 0 10, width=.2425\linewidth]{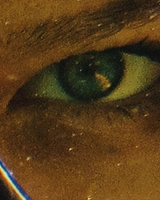}
    \put(4,4){\begin{color}{white}ground-truth\end{color}}
    \end{overpic}
    \end{minipage}
    
     \caption{JPEG artifact removal (quality factor $q=20$) example results. For each shown result we indicate the weight value for the perceptual ($L_1$, $L_2$, CTXDP, CTXL2, PDL) loss term. }
    \label{fig:jpeg-q20}
\end{figure}

We also evaluate the performance of the proposed loss in JPEG artifact removal and demosaicing under mild noise. Table~\ref{tab:jpeg-q20} summarizes the quantitative results of the different evaluated losses on JPEG artifact removal (quality factor $q=20$). Figure~\ref{fig:jpeg-q20} presents some selected results on JPEG artifact removal. The PDL loss successfully manages to restore the blockiness artifacts due to the sever JPEG compression, and recovers sharpness of the uncompressed image. 

\begin{figure}
    \centering
    \scriptsize

    \begin{overpic}[width=.24\linewidth]{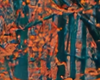}
    \put(2,2){\begin{color}{white}no-perceptual\end{color}}
    \end{overpic}
    \begin{overpic}[width=.24\linewidth]{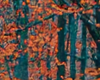}
    \put(2,2){\begin{color}{white}L1=0.01\end{color}}
    \end{overpic}
    \begin{overpic}[width=.24\linewidth]{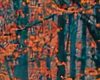}
    \put(2,2){\begin{color}{white}PDL=0.01\end{color}}
    \end{overpic}
    \begin{overpic}[width=.24\linewidth]{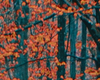}
    \put(2,2){\begin{color}{white}ground-truth\end{color}}
    \end{overpic}
    
    \vspace{.5em}
    
    \begin{overpic}[width=.24\linewidth]{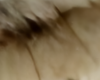}
    \put(2,2){\begin{color}{white}no-perceptual\end{color}}
    \end{overpic}
    \begin{overpic}[width=.24\linewidth]{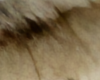}
    \put(2,2){\begin{color}{white}L1=0.01\end{color}}
    \end{overpic}
    \begin{overpic}[width=.24\linewidth]{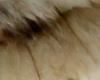}
    \put(2,2){\begin{color}{white}PDL=0.01\end{color}}
    \end{overpic}
    \begin{overpic}[width=.24\linewidth]{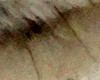}
    \put(2,2){\begin{color}{white}ground-truth\end{color}}
    \end{overpic}
    
    \vspace{.5em}

    \begin{overpic}[width=.24\linewidth]{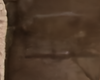}
    \put(2,2){\begin{color}{white}no-perceptual\end{color}}
    \end{overpic}
    \begin{overpic}[width=.24\linewidth]{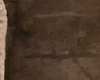}
    \put(2,2){\begin{color}{white}L1=0.01\end{color}}
    \end{overpic}
    \begin{overpic}[width=.24\linewidth]{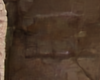}
    \put(2,2){\begin{color}{white}PDL=0.01\end{color}}
    \end{overpic}
    \begin{overpic}[width=.24\linewidth]{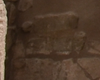}
    \put(2,2){\begin{color}{white}ground-truth\end{color}}
    \end{overpic}
    
    \vspace{.5em}
    
    \caption{Demosaicing under mild noise. For each shown result we indicate the weight value for the perceptual ($L_1$, PDL) loss term. }
    \label{fig:demosaic}
\end{figure}

In the case of demosaicing, the model is trained to predict an RGB image from the Bayer noisy mosaic. In Figure~\ref{fig:demosaic} we present a selection of results. Although the differences are subtle, the proposed PDL lead to images with better defined structures.

\begin{table}[h]
\renewcommand{\arraystretch}{0.8}
\setlength{\tabcolsep}{3.5pt}
\footnotesize
    \centering
    \begin{tabular}{lcccccc} \toprule
        Perceptual Loss	& PSNR	& MS-SSIM & LPIPS & NIQE & FID \\\midrule
        reference     &	-     &	-     &	-     & 3.166 & -  \\
        input	       &      29.54 &   0.953 & 0.208 & 4.267 & 28.24 \\
        no-perceptual      &  \textbf{31.77} &	\textbf{0.974} &	0.176 & 3.892 & 24.05 \\
     L1  ($\lambda=0.01$) &   31.70 &	0.973 &	0.144 & \textbf{3.622} & 17.94 \\
     L2  ($\lambda=0.01$) &   31.50 &	0.972 &	0.138 & 3.633 & 17.76 \\
   CTXDP ($\lambda=0.01$) &   31.48 &	0.972 &	0.113 & 3.812 & 16.20 \\
   CTXL2 ($\lambda=0.01$) &	  31.39 &	0.971 &	0.105 & 4.006 & 16.08  \\
     PDL ($\lambda=0.001$) &  31.63 &	0.973 &	0.135 & 3.687 & 16.40  \\
     PDL ($\lambda=0.005$) &  31.24 &	0.970 &	0.106 & 3.848 & \textbf{14.55} \\
     PDL ($\lambda=0.01$) &   31.07 &	0.969 &	\textbf{0.103} & 3.991 & 15.03 \\\midrule
    \end{tabular}
 \caption{JPEG artifact removal ($q=20$). All models were trained using the same model configuration and optimization parameters. All perceptual looses are computed on VGG16-conv4 features.}    \label{tab:jpeg-q20}
\end{table}

\subsection{Evaluation of the impact of the Feature Projection}
The proposed projected distribution loss requires to define a set of projection directions. In this section we present some results for adopting different projection schemes. We tested different alternatives for computing the projected distributions. Projection on each input feature which amounts to comparing the features marginal distributions (Id), a random combination of two randomly picked activation maps (R2P), a small random perturbation of the Identity matrix (RPP), and finally a random sampling on the sphere (RSP) -- which corresponds to a numerical approximation of the sliced Wasserstein distance. Additionally we evaluated the impact of the number of projections, as a factor of the original number of features. 
Table~\ref{tab:projection} summarizes the results. None of the tested alternatives seem to perform better than directly comparing the extracted features independently (Id). In general, all tested configurations performed similarly. We should note that VGG16 features are not normalized and therefore different features can have very different ranges. Normalizing them can lead to distortions in the geometry of the space and thus leading to artificially changing the relative importance of each feature. Recently~\cite{tarig2020why} showed that not all features are equally important. Finding better features or combinations of features will be a subject of our future work.

\begin{table}[t]
\renewcommand{\arraystretch}{0.8}
\footnotesize
    \centering
    \begin{tabular}{lccc} \toprule
Projection Type	& PSNR &	MS=SIM & LPIPS \\ \midrule
Id ($\lambda=0.001$) &	\textbf{27.10} & \textbf{0.906} &	0.250 \\
Id ($\lambda=0.005$) &	26.74 &	0.899 &	0.243 \\
\textbf{Id ($\boldsymbol{\lambda=0.01}$)} &	   26.62 & 0.898 & \textbf{0.233} \\
Id ($\lambda=0.01$) &	    26.36 &	0.893 &	0.239 \\\midrule
R2P-$8\times$ ($\lambda=0.001$) & 26.99 &	0.903 &	\textbf{0.255} \\
R2P-$4\times$ ($\lambda=0.005$) & 26.67 &	0.898 &	0.243 \\
R2P-$8\times$ ($\lambda=0.01$) & 26.45 &	0.895 &	0.239 \\
R2P-$8\times$ ($\lambda=0.10$) & 25.89	& 0.885 &	0.245 \\\midrule
RPP-$2\times$ ($\lambda=0.001$)	& 26.80	& 0.901 &	0.248 \\
RPP-$2\times$ ($\lambda=0.002$)	& 26.80	& 0.900 &	0.248 \\
RPP-$2\times$ ($\lambda=0.01$)  & 26.41	& 0.894 &	0.245 \\\midrule
RSP-$1\times$ ($\lambda=0.001$) 	& 26.50	& 0.892 &	0.243 \\ 
RSP-$4\times$ ($\lambda=0.001$) 	& 26.98	& 0.904 &	0.249 \\
RSP-$4\times$ ($\lambda=0.005$) 	& 26.47	& 0.894 &	0.247 \\
RSP-$1\times$ ($\lambda=0.01$) 	& 25.75	& 0.880 &	0.250 \\\midrule
    \end{tabular}
    \vspace{.2em}
    \caption{Evaluation of the impact of the feature projection.
     The different projection alternatives evaluated are:  comparing the original features marginal distributions independently (Id), a random combination of two randomly picked features (R2P), a small random perturbation of the original feature (RPP), and finally random projections on any possible direction on the sphere (RSP) -- which corresponds to a numerical approximation of the sliced Wasserstein distance. Additionally, we evaluated the impact of the number of projections, as a factor of the original number of features (-$n\times$, with $n=1, 2, 4, 8$). The best results are highlighted in bold.}
    \label{tab:projection}
\end{table}

\section{Conclusion}
\label{sec:discussion}

We have presented an alternative perceptual loss that can be used to train any image restoration model. The proposed projected distribution loss is straightforward to implement being based on comparing 1D marginal distributions. Our loss leads to superior or similar results on all the five tested image restoration applications. In this work we focused on comparing CNN features that have been shown to capture relevant characteristics to visual perception. As a future work we would like to investigate alternative schemes to project the multidimensional feature space into lower dimensions where the proposed distribution loss can be applied.

\bibliographystyle{IEEEtran}
\bibliography{egbib}

\begin{thebibliography}{10}
\providecommand{\url}[1]{#1}
\csname url@samestyle\endcsname
\providecommand{\newblock}{\relax}
\providecommand{\bibinfo}[2]{#2}
\providecommand{\BIBentrySTDinterwordspacing}{\spaceskip=0pt\relax}
\providecommand{\BIBentryALTinterwordstretchfactor}{4}
\providecommand{\BIBentryALTinterwordspacing}{\spaceskip=\fontdimen2\font plus
\BIBentryALTinterwordstretchfactor\fontdimen3\font minus
  \fontdimen4\font\relax}
\providecommand{\BIBforeignlanguage}[2]{{%
\expandafter\ifx\csname l@#1\endcsname\relax
\typeout{** WARNING: IEEEtran.bst: No hyphenation pattern has been}%
\typeout{** loaded for the language `#1'. Using the pattern for}%
\typeout{** the default language instead.}%
\else
\language=\csname l@#1\endcsname
\fi
#2}}
\providecommand{\BIBdecl}{\relax}
\BIBdecl

\bibitem{zhao2016loss}
H.~Zhao, O.~Gallo, I.~Frosio, and J.~Kautz, ``Loss functions for image
  restoration with neural networks,'' \emph{IEEE Transactions on computational
  imaging}, vol.~3, no.~1, pp. 47--57, 2016.

\bibitem{lim2017enhanced}
B.~Lim, S.~Son, H.~Kim, S.~Nah, and K.~Mu~Lee, ``Enhanced deep residual
  networks for single image super-resolution,'' in \emph{IEEE conference on
  computer vision and pattern recognition workshops}, 2017, pp. 136--144.

\bibitem{tao2018scale}
X.~Tao, H.~Gao, X.~Shen, J.~Wang, and J.~Jia, ``Scale-recurrent network for
  deep image deblurring,'' in \emph{IEEE Conference on Computer Vision and
  Pattern Recognition}, 2018, pp. 8174--8182.

\bibitem{chen2018learning}
C.~Chen, Q.~Chen, J.~Xu, and V.~Koltun, ``Learning to see in the dark,'' in
  \emph{IEEE Conference on Computer Vision and Pattern Recognition}, 2018, pp.
  3291--3300.

\bibitem{goodfellow2014generative}
I.~Goodfellow, J.~Pouget-Abadie, M.~Mirza, B.~Xu, D.~Warde-Farley, S.~Ozair,
  A.~Courville, and Y.~Bengio, ``Generative adversarial nets,'' in
  \emph{Advances in neural information processing systems}, 2014, pp.
  2672--2680.

\bibitem{arjovsky2017wasserstein}
M.~Arjovsky, S.~Chintala, and L.~Bottou, ``Wasserstein gan,'' \emph{arXiv
  preprint arXiv:1701.07875}, 2017.

\bibitem{ledig2017photo}
C.~Ledig, L.~Theis, F.~Husz{\'a}r, J.~Caballero, A.~Cunningham, A.~Acosta,
  A.~Aitken, A.~Tejani, J.~Totz, Z.~Wang \emph{et~al.}, ``Photo-realistic
  single image super-resolution using a generative adversarial network,'' in
  \emph{IEEE conference on computer vision and pattern recognition}, 2017, pp.
  4681--4690.

\bibitem{isola2017image}
P.~Isola, J.-Y. Zhu, T.~Zhou, and A.~A. Efros, ``Image-to-image translation
  with conditional adversarial networks,'' in \emph{IEEE conference on computer
  vision and pattern recognition}, 2017, pp. 1125--1134.

\bibitem{kupyn2018deblurgan}
O.~Kupyn, V.~Budzan, M.~Mykhailych, D.~Mishkin, and J.~Matas, ``Deblurgan:
  Blind motion deblurring using conditional adversarial networks,'' in
  \emph{IEEE conference on computer vision and pattern recognition}, 2018, pp.
  8183--8192.

\bibitem{kupyn2019deblurgan}
O.~Kupyn, T.~Martyniuk, J.~Wu, and Z.~Wang, ``Deblurgan-v2: Deblurring
  (orders-of-magnitude) faster and better,'' in \emph{IEEE International
  Conference on Computer Vision}, 2019, pp. 8878--8887.

\bibitem{arora2017generalization}
S.~Arora, R.~Ge, Y.~Liang, T.~Ma, and Y.~Zhang, ``Generalization and
  equilibrium in generative adversarial nets (gans),'' in \emph{International
  Conference on Machine Learning}, 2017, pp. 224--232.

\bibitem{salimans2016improved}
T.~Salimans, I.~Goodfellow, W.~Zaremba, V.~Cheung, A.~Radford, and X.~Chen,
  ``Improved techniques for training gans,'' in \emph{Advances in neural
  information processing systems}, 2016, pp. 2234--2242.

\bibitem{cohen2018distribution}
J.~P. Cohen, M.~Luck, and S.~Honari, ``Distribution matching losses can
  hallucinate features in medical image translation,'' in \emph{International
  conference on medical image computing and computer-assisted
  intervention}.\hskip 1em plus 0.5em minus 0.4em\relax Springer, 2018, pp.
  529--536.

\bibitem{simonyan2014very}
K.~Simonyan and A.~Zisserman, ``Very deep convolutional networks for
  large-scale image recognition,'' \emph{arXiv preprint arXiv:1409.1556}, 2014.

\bibitem{johnson2016perceptual}
J.~Johnson, A.~Alahi, and L.~Fei-Fei, ``Perceptual losses for real-time style
  transfer and super-resolution,'' in \emph{European conference on computer
  vision}.\hskip 1em plus 0.5em minus 0.4em\relax Springer, 2016, pp. 694--711.

\bibitem{zhang2018unreasonable}
R.~Zhang, P.~Isola, A.~A. Efros, E.~Shechtman, and O.~Wang, ``The unreasonable
  effectiveness of deep features as a perceptual metric,'' in \emph{IEEE
  conference on computer vision and pattern recognition}, 2018, pp. 586--595.

\bibitem{tarig2020why}
T.~Tariq, O.~T. Tursun, M.~Kim, and P.~Didyk, ``Why are deep representations
  good perceptual quality features?'' in \emph{Computer Vision -- ECCV 2020},
  A.~Vedaldi, H.~Bischof, T.~Brox, and J.-M. Frahm, Eds.\hskip 1em plus 0.5em
  minus 0.4em\relax Cham: Springer International Publishing, 2020, pp.
  445--461.

\bibitem{mechrez2018contextual}
R.~Mechrez, I.~Talmi, and L.~Zelnik-Manor, ``The contextual loss for image
  transformation with non-aligned data,'' in \emph{European Conference on
  Computer Vision (ECCV)}, 2018, pp. 768--783.

\bibitem{mechrez2018maintaining}
R.~Mechrez, I.~Talmi, F.~Shama, and L.~Zelnik-Manor, ``Maintaining natural
  image statistics with the contextual loss,'' in \emph{Asian Conference on
  Computer Vision}.\hskip 1em plus 0.5em minus 0.4em\relax Springer, 2018, pp.
  427--443.

\bibitem{zhang2019zoom}
X.~Zhang, Q.~Chen, R.~Ng, and V.~Koltun, ``Zoom to learn, learn to zoom,'' in
  \emph{IEEE Conference on Computer Vision and Pattern Recognition}, 2019, pp.
  3762--3770.

\bibitem{gatys2016image}
L.~A. Gatys, A.~S. Ecker, and M.~Bethge, ``Image style transfer using
  convolutional neural networks,'' in \emph{IEEE conference on computer vision
  and pattern recognition}, 2016, pp. 2414--2423.

\bibitem{li2017demystifying}
Y.~Li, N.~Wang, J.~Liu, and X.~Hou, ``Demystifying neural style transfer,'' in
  \emph{26th International Joint Conference on Artificial Intelligence}, ser.
  IJCAI'17.\hskip 1em plus 0.5em minus 0.4em\relax AAAI Press, 2017, p.
  2230–2236.

\bibitem{li2017universal}
Y.~Li, C.~Fang, J.~Yang, Z.~Wang, X.~Lu, and M.-H. Yang, ``Universal style
  transfer via feature transforms,'' in \emph{Advances in neural information
  processing systems}, 2017, pp. 386--396.

\bibitem{villani2008optimal}
C.~Villani, \emph{Optimal transport: old and new}.\hskip 1em plus 0.5em minus
  0.4em\relax Springer Science \& Business Media, 2008, vol. 338.

\bibitem{peyre2019computational}
G.~Peyr{\'e}, M.~Cuturi \emph{et~al.}, ``Computational optimal transport: With
  applications to data science,'' \emph{Foundations and Trends{\textregistered}
  in Machine Learning}, vol.~11, no. 5-6, pp. 355--607, 2019.

\bibitem{cuturi2013sinkhorn}
M.~Cuturi, ``Sinkhorn distances: Lightspeed computation of optimal transport,''
  in \emph{Advances in neural information processing systems}, 2013, pp.
  2292--2300.

\bibitem{rabin2011wasserstein}
J.~Rabin, G.~Peyr{\'e}, J.~Delon, and M.~Bernot, ``Wasserstein barycenter and
  its application to texture mixing,'' in \emph{International Conference on
  Scale Space and Variational Methods in Computer Vision}.\hskip 1em plus 0.5em
  minus 0.4em\relax Springer, 2011, pp. 435--446.

\bibitem{dong2015image}
C.~Dong, C.~C. Loy, K.~He, and X.~Tang, ``Image super-resolution using deep
  convolutional networks,'' \emph{IEEE transactions on pattern analysis and
  machine intelligence}, vol.~38, no.~2, pp. 295--307, 2015.

\bibitem{zhao2017loss}
H.~{Zhao}, O.~{Gallo}, I.~{Frosio}, and J.~{Kautz}, ``Loss functions for image
  restoration with neural networks,'' \emph{IEEE Transactions on Computational
  Imaging}, vol.~3, no.~1, pp. 47--57, 2017.

\bibitem{talebi2018learned}
H.~Talebi and P.~Milanfar, ``Learned perceptual image enhancement,'' in
  \emph{2018 IEEE international conference on computational photography
  (ICCP)}.\hskip 1em plus 0.5em minus 0.4em\relax IEEE, 2018, pp. 1--13.

\bibitem{talebi2018nima}
------, ``Nima: Neural image assessment,'' \emph{IEEE Transactions on Image
  Processing}, vol.~27, no.~8, pp. 3998--4011, 2018.

\bibitem{blau2018perception}
Y.~Blau and T.~Michaeli, ``The perception-distortion tradeoff,'' in \emph{IEEE
  Conference on Computer Vision and Pattern Recognition}, 2018, pp. 6228--6237.

\bibitem{frogner2015learning}
C.~Frogner, C.~Zhang, H.~Mobahi, M.~Araya, and T.~A. Poggio, ``Learning with a
  wasserstein loss,'' in \emph{Advances in neural information processing
  systems}, 2015, pp. 2053--2061.

\bibitem{wu2019sliced}
J.~Wu, Z.~Huang, D.~Acharya, W.~Li, J.~Thoma, D.~P. Paudel, and L.~V. Gool,
  ``Sliced wasserstein generative models,'' in \emph{IEEE conference on
  computer vision and pattern recognition}, 2019, pp. 3713--3722.

\bibitem{zhang2020deepemd}
C.~Zhang, Y.~Cai, G.~Lin, and C.~Shen, ``Deepemd: Few-shot image classification
  with differentiable earth mover's distance and structured classifiers,'' in
  \emph{IEEE/CVF Conference on Computer Vision and Pattern Recognition}, 2020,
  pp. 12\,203--12\,213.

\bibitem{deza2006dictionary}
M.-M. Deza and E.~Deza, \emph{Dictionary of distances}.\hskip 1em plus 0.5em
  minus 0.4em\relax Elsevier, 2006.

\bibitem{bonneel2015sliced}
N.~Bonneel, J.~Rabin, G.~Peyr{\'e}, and H.~Pfister, ``Sliced and radon
  wasserstein barycenters of measures,'' \emph{Journal of Mathematical Imaging
  and Vision}, vol.~51, no.~1, pp. 22--45, 2015.

\bibitem{kolouri2019generalized}
S.~Kolouri, K.~Nadjahi, U.~Simsekli, R.~Badeau, and G.~Rohde, ``Generalized
  sliced wasserstein distances,'' in \emph{Advances in Neural Information
  Processing Systems}, 2019, pp. 261--272.

\bibitem{deshpande2018generative}
I.~Deshpande, Z.~Zhang, and A.~G. Schwing, ``Generative modeling using the
  sliced wasserstein distance,'' in \emph{IEEE conference on computer vision
  and pattern recognition}, 2018, pp. 3483--3491.

\bibitem{deshpande2019max}
I.~Deshpande, Y.-T. Hu, R.~Sun, A.~Pyrros, N.~Siddiqui, S.~Koyejo, Z.~Zhao,
  D.~Forsyth, and A.~G. Schwing, ``Max-sliced wasserstein distance and its use
  for gans,'' in \emph{IEEE conference on computer vision and pattern
  recognition}, 2019, pp. 10\,648--10\,656.

\bibitem{agustsson2017ntire}
E.~Agustsson and R.~Timofte, ``Ntire 2017 challenge on single image
  super-resolution: Dataset and study,'' in \emph{IEEE Conference on Computer
  Vision and Pattern Recognition Workshops}, 2017, pp. 126--135.

\bibitem{he2016deep}
K.~He, X.~Zhang, S.~Ren, and J.~Sun, ``Deep residual learning for image
  recognition,'' in \emph{IEEE conference on computer vision and pattern
  recognition}, 2016, pp. 770--778.

\bibitem{nah2017deep}
S.~Nah, T.~Hyun~Kim, and K.~Mu~Lee, ``Deep multi-scale convolutional neural
  network for dynamic scene deblurring,'' in \emph{IEEE Conference on Computer
  Vision and Pattern Recognition}, 2017, pp. 3883--3891.

\bibitem{ronneberger2015u}
O.~Ronneberger, P.~Fischer, and T.~Brox, ``U-net: Convolutional networks for
  biomedical image segmentation,'' in \emph{International Conference on Medical
  image computing and computer-assisted intervention}.\hskip 1em plus 0.5em
  minus 0.4em\relax Springer, 2015, pp. 234--241.

\bibitem{mao2016image}
X.~Mao, C.~Shen, and Y.-B. Yang, ``Image restoration using very deep
  convolutional encoder-decoder networks with symmetric skip connections,'' in
  \emph{Advances in neural information processing systems}, 2016, pp.
  2802--2810.

\bibitem{gao2019dynamic}
H.~Gao, X.~Tao, X.~Shen, and J.~Jia, ``Dynamic scene deblurring with parameter
  selective sharing and nested skip connections,'' in \emph{IEEE Conference on
  Computer Vision and Pattern Recognition}, 2019, pp. 3848--3856.

\bibitem{wang2003multiscale}
Z.~Wang, E.~P. Simoncelli, and A.~C. Bovik, ``Multiscale structural similarity
  for image quality assessment,'' in \emph{The Thrity-Seventh Asilomar
  Conference on Signals, Systems \& Computers, 2003}, vol.~2.\hskip 1em plus
  0.5em minus 0.4em\relax Ieee, 2003, pp. 1398--1402.

\bibitem{delbracio2015burst}
M.~Delbracio and G.~Sapiro, ``Burst deblurring: Removing camera shake through
  fourier burst accumulation,'' in \emph{IEEE Conference on Computer Vision and
  Pattern Recognition}, 2015, pp. 2385--2393.

\end{thebibliography}

\ifpeerreview \else

\begin{IEEEbiography}[{\includegraphics[width=1in,height=1.25in,clip,keepaspectratio]{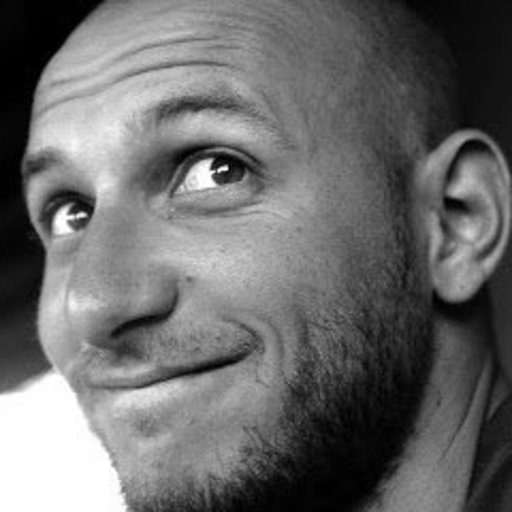}}]{Mauricio Delbracio} is a research scientist at Google Research. Before joining Google in 2019, he was an Assistant Professor at the Department of Electrical Engineering, Universidad de la Rep\'ublica (UdelaR), Uruguay. From 2013 to 2016 he was a postdoctoral researcher with the ECE Department at Duke University. He received the B.Sc degree in electrical engineering from UdelaR, Montevideo, in 2006, and the M.Sc. and Ph.D. degrees in applied mathematics from \'Ecole Normale Sup\'erieure de Cachan (ENS-Cachan), France, in 2009 and 2013 respectively. 
His current research focuses on algorithms, data analysis and applications of machine learning to image and signal processing. In 2016 he was awarded the Early Career Prize from the Society for Industrial and Applied Mathematics (SIAM) Activity Group on Imaging Science in 2016 for his important contributions to image processing.
\end{IEEEbiography}

\begin{IEEEbiography}[{\includegraphics[width=1in,height=1.25in,clip,keepaspectratio]{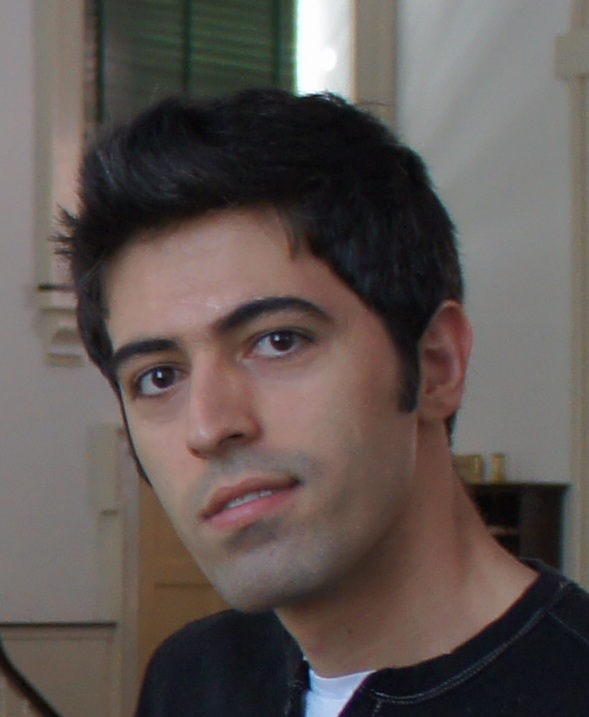}}]{Hossein Talebi} received the B.S. and M.S. degrees in electrical engineering from the Isfahan University of Technology, Iran, and the Ph.D. degree in electrical engineering from the University of California at Santa Cruz, Santa Cruz, CA, USA. Since 2015, he has been with Google Research, Mountain View, CA, where he works on computational imaging, image processing and machine learning problems.

\end{IEEEbiography}

\begin{IEEEbiography}[{\includegraphics[width=1in,height=1.25in,clip,keepaspectratio]{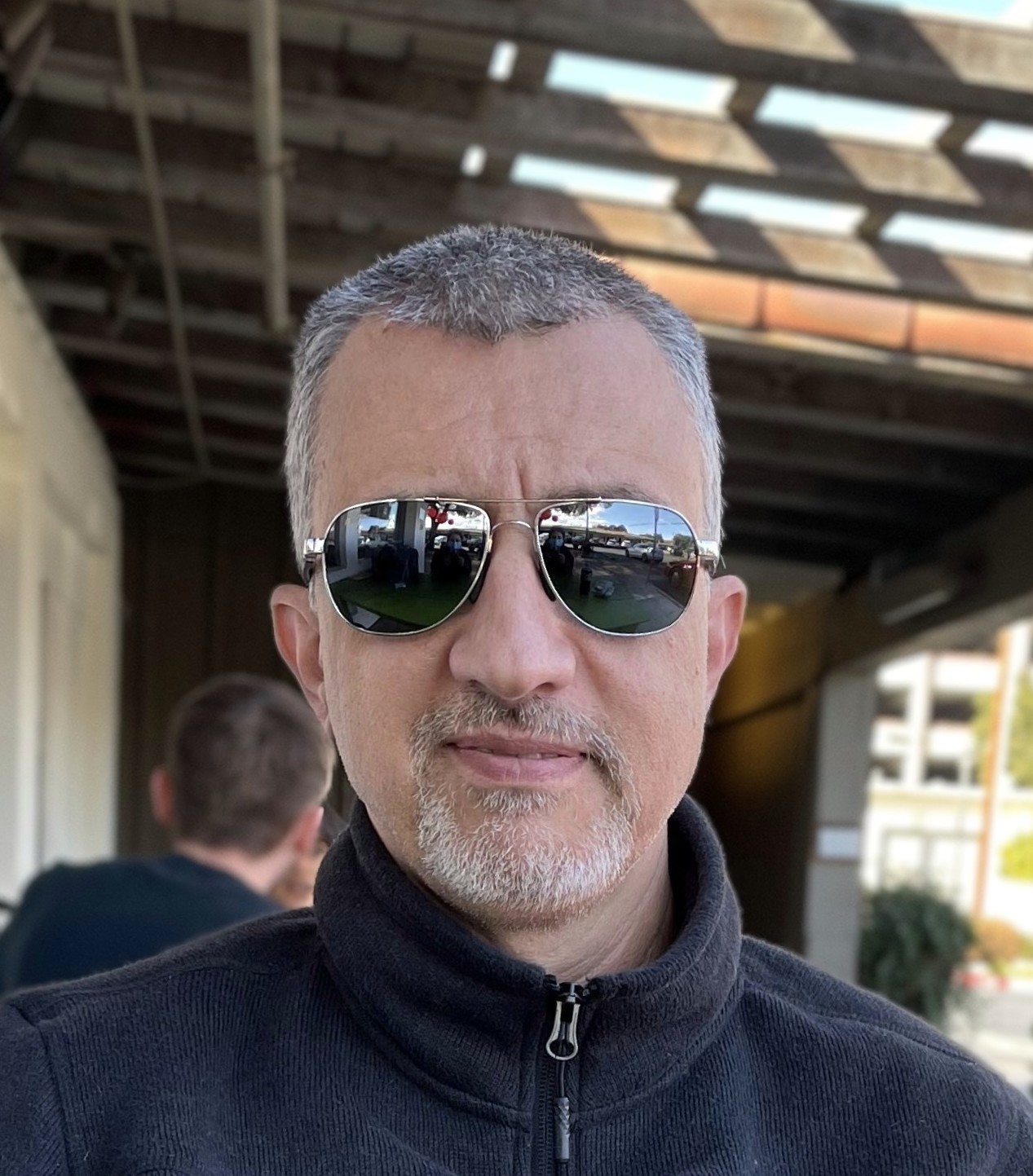}}]{Peyman Milanfar} is a Principal Scientist / Director at Google Research, where he leads the Computational Imaging team. Prior to this, he was a Professor of Electrical Engineering at UC Santa Cruz from 1999-2014. He was Associate Dean for Research at the School of Engineering from 2010-12. From 2012-2014 he was on leave at Google-x, where he helped develop the imaging pipeline for Google Glass. 
Most recently, Peyman's team at Google developed the digital zoom pipeline for the Pixel phones, which includes the multi-frame super-resolution (Super Res Zoom) pipeline (blog, and project website), and the RAISR upscaling algorithm.  In addition, the Night Sight mode on Pixel 3 uses our Super Res Zoom technology to merge images (whether you zoom or not) for vivid shots in low light, including astrophotography.
Peyman received his undergraduate education in electrical engineering and mathematics from the University of California, Berkeley, and the MS and PhD degrees in electrical engineering from the Massachusetts Institute of Technology. He holds 15 patents, several of which are commercially licensed. He founded MotionDSP, which was acquired by Cubic Inc. (NYSE:CUB). 
Peyman has been keynote speaker at numerous technical conferences including Picture Coding Symposium (PCS), SIAM Imaging Sciences, SPIE, and the International Conference on Multimedia (ICME). Along with his students, he has won several best paper awards from the IEEE Signal Processing Society. 
He is a Distinguished Lecturer of the IEEE Signal Processing Society, and a Fellow of the IEEE ``for contributions to inverse problems and super-resolution in imaging.''
\end{IEEEbiography}

\fi

\end{document}